\documentclass[12pt]{supplementary}  %

\usepackage[utf8]{inputenc} %
\usepackage{csquotes} %
\usepackage{booktabs}       %
\usepackage{nicefrac}       %
\usepackage{microtype}      %
\usepackage{times}
\usepackage{todonotes}
\usepackage{wrapfig}
\usepackage{cleveref}
\usepackage{supertabular}
\usepackage{multicol}
\usepackage{multirow}
\usepackage{ulem}
\usepackage{xcolor}
\usepackage[export]{adjustbox}
\usepackage{mymath}
\usepackage{float}
\usepackage{algorithm}
\usepackage{algpseudocode}

\newcommand{\vizdoom}{ViZDoom}
\newcommand{\aithor}{AI2-THOR}
\newcommand{\brax}{Brax}

\newcommand{\tosupplementary}{For a detailed overview of each experiment, please refer to the appendices.}

\title{Filter-Aware Model-Predictive Control}

\author{%
	\Name{Bar\i\c{s} Kayal\i bay} \Email{bkayalibay@argmax.ai}\\
 \Name{Atanas Mirchev} \Email{atanas.mirchev@argmax.ai}\\
 \Name{Ahmed Agha} \Email{ahmed.agha@argmax.ai}\\
 \Name{Patrick {van der Smagt}}\\
 \Name{Justin Bayer} \Email{bayerj@argmax.ai}\\
 \addr Machine Learning Research Lab, Volkswagen Group, Munich%
}

\begin{document}

\maketitle

\begin{abstract}
Partially-observable problems pose a trade-off between reducing costs and gathering information.
They can be solved optimally by planning in belief space, but that is often prohibitively expensive.
Model-predictive control (MPC) takes the alternative approach of using a state estimator to form a belief over the state, and then plan in state space.
This ignores potential future observations during planning and, as a result, cannot actively increase or preserve the certainty of its own state estimate.
We find a middle-ground between planning in belief space and completely ignoring its dynamics by only reasoning about its {\it future accuracy}.
Our approach, filter-aware MPC, penalises the loss of information by what we call “trackability”, the expected error of the state estimator.
We show that model-based simulation allows condensing trackability into a neural network, which allows fast planning.
In experiments involving visual navigation, realistic every-day environments and a two-link robot arm, we show that filter-aware MPC vastly improves regular MPC.
 \end{abstract}

\section{Introduction}
In partially observable Markov decision problems (POMDPs) we can only estimate the system state from a history of observations.
This makes planning in POMDPs complicated, as the planner must find an optimal balance between reducing costs according to the current state estimate, and gathering new observations which will refine that state estimate, allowing for even better cost minimisation.

Optimally trading-off cost reduction and information gathering is only possible by planning in the space of beliefs, but that is often too expensive.
Another approach, taken by model-predictive control (MPC),\footnote{Related terms are receding horizon control, limited lookahead control, and rollout. We use the term MPC and clarify what we mean in \cref{sec:background}.} is to decouple state estimation and planning completely.
The planner only focuses on costs, finding the best sequence of controls starting from the current belief.
Planning in this way is as cheap as planning in a fully observable problem, but allows no way of actively acquiring new information or preserving the information we already have.

We try to find a middle ground between these two approaches.
In our approach, we do not plan in belief space, but we also do not ignore future beliefs completely.
Instead, we settle for reasoning about the future accuracy of the state estimate.
This results in a controller that can actively maintain a desired amount of accuracy in its belief.

As a motivating example, we can look at visual navigation, where a visual tracker uses RGB-D images to track the location of a robot which is trying to reach some target coordinates.
Following the shortest path might force the robot to pass through a long corridor that does not contain enough information for the visual tracker to work with.
The state estimate might degrade so much that it is of no help to the planner.
If we have a way of predicting the tracker's accuracy around this corridor, we can inform the planner about it, which can then decide to follow a slightly longer path, but one where the state estimate will remain accurate. 

We propose a straightforward method for blending this aspect into MPC.
We introduce the idea of trackability, which maps a system's state to the expected error of a state estimator.
By putting a threshold on trackability, we add an additional constraint to the MPC problem.
Solving this augmented problem minimises the total cost of the original control task, while maintaining a desired degree of accuracy in the state estimator.
We show that trackability can be seen as a value function of a new MDP where the state contains both the original system state and the state estimator's belief.
This allows us to learn a neural network that predicts trackability using standard approximate dynamic programming techniques.
We verify these ideas in experiments designed to illustrate the problem, including visual navigation, realistic every-day environments and a two-link robot arm.
In all cases, we obtain substantial improvements over baselines that ignore how the belief will evolve over time.

Our exact contributions are:
\begin{itemize}
\item We formulate a measure of how accurate state estimation will be under a plan and a starting state. We learn this measure with a neural network for fast planning.
\item We present filter-aware MPC, a version of MPC that can prevent state estimation errors by putting a threshold on trackability.
\item We empirically show that filter-aware MPC improves regular MPC and that it performs competitively against other baselines using observations.
\end{itemize}

\section{Related work}

\citet{aastrom1965optimal} was among the first to publish theoretical results on decision-making with missing information.
\citet{sondik1971optimal} developed a method for finite-horizon optimal control, showing that the value function is piecewise-linear in this setting.
Later works dealt with the infinite horizon case \citep{sawaki1978optimal,sondik1978optimal}.
A large body of work focused on performance improvements for exact inference \citep{Littman96algorithmsfor, KaelblingLC98, cassandra1998incremental}.
Others sought to find efficient approximations \citep{parr1995approximating,thrun1999Monte,Zhang_2001,pineau2003point}.
Recent works presented tree search algorithms using particle filters \citep{Sunberg2017Online,sunberg2017Value}, allowing to solve POMDPs with continuous states, controls and observations.

Many works focused on the case with Gaussian beliefs.
These include \citep{platt2010belief}, working with extended Kalman filter updates and \citep{todorov2005stochastic}, which allows belief planning in an LQG setting with multiplicative noise.
\citet{vanDenBerg2021Efficient} presented a value learning algorithm with the EKF.
\citet{rafieisakhaei2017tLQG} improved LQG planning by optimising for the lowest perception uncertainty.
\citet{rahman2020Uncertainty} introduced a vision-oriented approach with Gaussian beliefs where the uncertainty of the belief is constrained.
Similar measures were taken in \citep{Hovd2005Interaction} and \citep{Bohm2008Avoidance}.

Our approach in this paper is also related to {\it coastal navigation} \citep{roy1999coastal, Roy1999CoastalNR}, which focuses on preserving the accuracy of the state estimate, rather than full planning in belief space.
Similarly, {\it belief roadmaps} \citep{prentice2008belief} try to find a path from a start location to a target while minimising the uncertainty of an EKF.
This was later extended to work with unscented Kalman filters \citep{he2008planning}.
Belief roadmaps were built upon by \citet{zheng2021Belief} by constraining the covariance of the belief.

Our method differs from these works in a number of ways.
First, we look at the expected error of a state estimator.
The expected error will be essentially the same as the entropy if the state estimate is the true posterior (or an accurate approximation), but is also applicable to state estimators that are not accurate approximations of the true posterior in all cases and ones that produce a point estimate.
Second, we limit expected tracking errors into the future (beyond the planning horizon) and rely on approximate dynamic programming techniques for that, which is an aspect we have not found in the literature.
Finally, since our approach only operates in state space, we are able to work on high-dimensional problems where (approximate) belief planning is typically infeasible, e.g. problems with image observations.

Several recent papers showed great results on partially observable problems using model-based methods \citep{karl2017unsupervised, Hafner2019learning, becker2020learning, hafner2021mastering}.
These works learn a policy that works with full state information by simulating the problem with a generative model.
At test time, an inference network maps the observation stream to a latent state, which is passed to the policy.
This strategy is similar to MPC with a state estimator: the policy takes the current state estimate and tries to find an optimal strategy.
The work of \citet{IglZLWW18} and \citet{HanDT20} are notable exceptions, as they train policies in belief space.
Similarly, \citet{lee2019stochastic} have presented a latent space algorithm which approximates belief planning.
Our work mainly differs from these in that we focus on MPC, which has desirable properties such as state-constraints, stability and performance guarantees.

Finally, we would like to point the reader to the field of dual control \citep{mesbah2018Stochastic}, which contains methods building on MPC that deal with the trade-off between cost reduction and state estimation \citep{kohler2021robust}.
\section{Background}\label{sec:background}
\subsection{Systems with Imperfect State Information}
We work with problems where a hidden state $\state_t \in \mathbb{R}^{N_\state}$ changes over time according to some transition function $f$ given controls $\control_t  \in \mathbb{R}^{N_\control}$ and process noise $\noise_t$.
The hidden state can only be estimated through observations $\obs_t  \in \mathbb{R}^{N_\obs}$ that are produced by an emission function $g$ subject to measurement noise variables $\obsnoise_t$.
The process and the observation noise might be state and control dependent with known distributions and dependency.
Using some control $\control_t$ in some state $\state_t$ leads to a cost $\cost_t \in \mathbb{R}^+$ which is determined by the cost function $c$.
This setup can be summarised:
\begin{gather}
	\state_{t+1} = f(\state_t, \control_t, \noise_t) \quad \cost_t= c(\state_t, \control_t) \quad \obs_t = g(\state_t, \obsnoise_t),
\end{gather}
with initial state distribution $p(\state_1)$.
The controls are given by a stationary policy $\control_t = \policy(\obs_{1:t}, \control_{1:t-1})$ that is conditioned on all previous observations and controls.
The expected total cost of a policy $\policy$ is:
\eq{
\totalcost{\pi}(\state) = \mathbb{E}\Big[\sum_{t=1}^\infty \discount^{t-1} c(\state_t, \control_t)\mid \state_1 = \state\Big],
}
where $\discount \in (0, 1)$.
An optimal policy attains minimal expected cost, i.e. $\optimalpolicy = \arg \min_\pi \mathbb{E} \left [ J^\pi (\state) \right ]$.

\subsection{Model Predictive Control for Imperfect State Information}
A variant of MPC related to open-loop feedback control \citep{dreyfus1965dynamic} approximates this problem by devising an optimal plan for the first $K$ steps and using its first step:
\eq{\label{eq:mpc}
	\pi_{\text{MPC}}(\obs_{1:t}, \control_{1:t-1}) = \arg \min_{\control_t} \min_{\control_{t:t+K-1}} \mathbb{E}_{\state_t|\obs_{1:t}, \control_{1:t-1}} \left [ \sum_{k=t}^{t+K-1} \discount^{k-t} c(\state_k, \control_k) + \discount^{K} {\apxtotalcost}(\state_{t+K}) \right ].
\numberthis
}
A terminal cost function $\apxtotalcost$ is used to alleviate some of the bias resulting from omitting the future.

\subsection{State Estimation for MPC}
Sampling the future states $\state_{t:t+K}$ in \cref{eq:mpc} necessitates a distribution over the current state, the {\it belief}.
This is done by a state estimator, which reads the observations into an internal state, or {\it carry}, which we denote with $\carry$.
The carry is updated by a function $h(\cdot, \cdot)$ which looks at the current observation and the previous carry and control: $\carry_{t+1}=h(\carry_t, \obs_{t+1}, \control_t)$.

The form of the carry depends on the state estimator.
For a particle filter, the carry would be the set of particles and weights.
For an RNN, it would be the hidden state vector.
Using the concept of the carry, we define the belief as $q(\state_t \mid \carry_t)$, which acts as the distribution to use in \cref{eq:mpc}.
It is often reasonable to assume that the initial state of the system is known (e.g.\ when we are only interested in relative motion).
We formalise this as setting $\carry_1:=\carryperf_\state$, where $\carry_1$ is the initial carry and $\carryperf_\state$ is a carry that deterministically identifies the true state $\state$.
We generally use the same notation $\state$ for both true and estimated states to keep notation simple.

\subsection{Suboptimality of MPC}
Model predictive control is suboptimal because it assumes the agent will collect no future observations.
In fact, it can be shown that this control scheme results from a single step of policy improvement over open-loop planning \citep{Bertsekas05} in the rollout framework.
While this often results in dramatic improvements, we argue that there remains a blind spot in MPC.

As an extreme case, consider an additional action which grants the agent access to the true state at the {\it next} step.
There is no incentive for the MPC-based policy to take it, as it will not reduce the expected cost of open-loop plans.

The last insight serves as the major motivation of our work: we will explicitly integrate the expected performance of a state estimator in the planning of MPC.
\section{Filter-Aware MPC}\label{sec:method}
We aim to improve MPC by letting it distinguish between actions not just in terms of how much they reduce the cost, but also how they influence the belief.
We want to find out the future accuracy of the state estimate under different plans and put a constraint on that value such that we only pick plans that guarantee a certain level of accuracy.

At every time step, any state estimator will introduce some inaccuracy, even the optimal Bayes filter.
We denote this error with $\errorf(\state_t, \carry_t)$. In the case of a probabilistic state estimator, this might be its negative log-likelihood $-\log q(\state_t~|~\carry_t)$, or the sum of squares in case of a deterministic one.
If we take the example of a Kalman filter, $\carry_t$ would be the mean and covariance matrix of the posterior distribution over the system state at time $t$, and $\errorf(\state_t, \carry_t)$ would be the negative log-likelihood of $\state_t$ under that distribution.

For any policy, the expected total state estimation error starting from a state $\state$ and a carry $\carry$ is:
\eq{\label{eq:trackerr}
	\track(\state, \carry) = \mathbb{E}\Big[\sum_{t=1}^\infty \discount^{t-1}\errorf(\state_t,\carry_t)\mid \state_1 = \state, \carry_1 =  \carry\Big],
	\numberthis
}
where the expectation is over both the transition and the emission noise and the controls are picked by some policy $\policy$.
Future tracking errors are discounted by $\discount \in (0, 1)$.

We define $\track(\state, \carry)$ as the trackability of $\state$ under the policy $\policy$ and the carry $\carry$.
Given a sequence of controls $\control_{t:t+K-1}$ and stochastic dynamics, the trackability of the state at time $t^\prime$ will be stochastic.
We extend \cref{eq:mpc} with a chance constraint \citep{farina2016stochastic}:
\eq{\label{eq:filteraware}
\min_{\control_{t:t+K-1}} \quad &\mathbb{E}_{\state \sim q(\state_t \mid \carry_t)}\Big[\obj(\state)\Big],\\
\hspace{1.5em}\textrm{s.t.} \quad & \text{Pr}\bigl(\track(\state_{t+i}, \carry_{t+i}) \le \delta\bigr) \ge \Delta \quad \text{for} \quad i=1,2,\dots,K. \numberthis
}
Here, $\delta$ is a threshold which defines how much tracker error we are willing to tolerate and $\Delta$ is a minimum probability of satisfying this condition that we wish to guarantee.
We refer to this version of MPC as filter-aware MPC.
Just like regular MPC, filter-aware MPC minimises the total cost.
Unlike regular MPC, it also considers how a sequence of actions will affect the state estimate, only allowing plans where a minimum degree of state estimation accuracy can be achieved with high enough probability.

Unfortunately, optimising \cref{eq:filteraware} will not bring us very far in practice.
The problem is that checking the constraint for time $t+i$ requires us to predict the future carry $\carry_{t+i}$.
This is as computationally expensive as planning in belief space.
In this version of filter-aware MPC, checking the constraints comes at a greater expense than evaluating the MPC objective.

We make an approximation to arrive at a feasible algorithm.
We approximate $\track(\state_t, \carry_t) \approx \track(\state_t, \carryperf_{\state_t})$.
In other terms, when we check the trackability of the state visited in time $t$, we only care about tracking errors that will happen in future time steps $t^\prime > t$ as if the process was started fresh from $\state_t$ with a perfect state estimate.
In the remainder, we drop the conditioning on $\carryperf_\state$ and set $\track(\state) := \track(\state, \carryperf_{\state})$ for brevity.
We are able to do so because $\carryperf_\state$ is a function of $\state$.

We will use a multi-layer perceptron (MLP) to predict $\track(\state)$.
An MLP is a natural choice for $\track(\state)$, since this can be seen as the negative value function (or cost-to-go) of an MDP where the cost is the tracking error and the MDP-state combines the system state $\state$ and the carry $\carry$, though the latter does not enter into our computations since we are only interested in the case where $\carry=\carryperf_\state$.
Taking this angle, we use TD($\lambda$) \citep{sutton1988learning} to learn an MLP $\phi$ that approximates trackability.
Next, by evaluating $\phi(\state_t)$, we are able to check the constraints without ever reasoning about future state estimates directly.
\subsection{Learning Trackability}
Using the state estimator and our models for the transition and emission $f$ and $g$, we can create rollouts of a regular MPC policy.
These rollouts have the form $\mathcal{D}=\{\state^i_{1:T}, \carry^i_{1:T}, \error^i_{1:T-1}\}_{i=1}^N$, where we denote the state estimation error at time $t$ by $\error_t := \errorf(\state_t, \carry_t)$.

Given $\mathcal{D}$, we can train the neural network by minimising:
\eq{\label{eq:critic}
\min_{\phi} \sum_{i}^N (\phi(\state_1^i) -\track(\state_1^i; \lambda, \phi))^2,\numberthis
}
where $\track(\state_1^i;\lambda, \phi)$ is the $\lambda$-return of $\state_1^i$ defined as:
\eq{
\track(\state_1^i;\lambda, \phi) &= (1-\lambda)\sum_{k=1}^{T-1} \lambda^{k-1}\track_k(\state_1^i;\phi), &
\track_k(\state_1^i;\phi) &= \sum_{t=1}^k \discount^{t-1}\error_t^i + \discount^k\phi(\state_{k+1}^i).
}

\subsection{A Practical Implementation of Filter-Aware MPC}
Once the trackability net $\phi$ has been learned, any optimisation algorithm that can handle constraints can be used to optimise \cref{eq:filteraware} by replacing $\track(\state_{t+i}, \carry_{t+i}) \approx \phi(\state_{t+i})$.
In our experiments we used simple random search for simplicity and speed.
Here we sample $K$ candidate plans $\control_{1:H-1}$ from a proposal distribution and take the one that performs best among those that satisfy constraints.
If no plan can satisfy the constraints, we pick the one with the minimal violation.
This crude algorithm does not guarantee constraint satisfaction and cannot handle chance constraints, but we have found it to work well for our problems.

\section{Experiments}
We experiment in settings with different levels of complexity and realism.
\tosupplementary

\subsection{Toy Scenario}
We start with a toy experiment that demonstrates our ideas in a simple setting.
\begin{figure}[t]
    \floatconts{fig:toy}{
        \caption{
            \vspace{-2em}
	    Toy scenario of a 2D particle aiming for the green region on the left starting from the right.
	    The grey circle is a "dark zone" with higher observation noise.
	    (a) A random walk. (b) Learned trackability. (c) Filter-aware vs vanilla MPC. (d) Success rates of vanilla MPC on an easy system with no grey zone, vanilla MPC, filter-aware MPC and an RNN policy.
        }
    }%
    {
        \centering
        \subfigure[][b]{
            \includegraphics[width=0.21\linewidth]{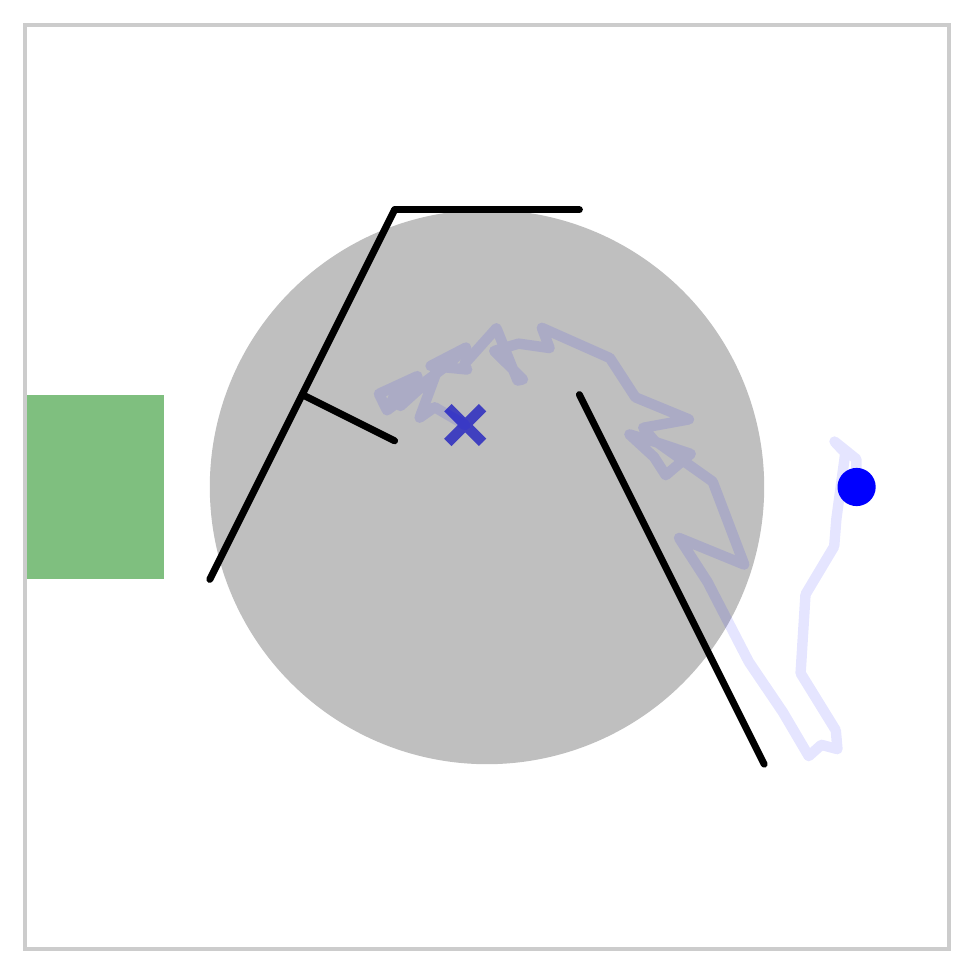}
        }\hfill
        \subfigure[][b]{
            \includegraphics[width=0.21\linewidth]{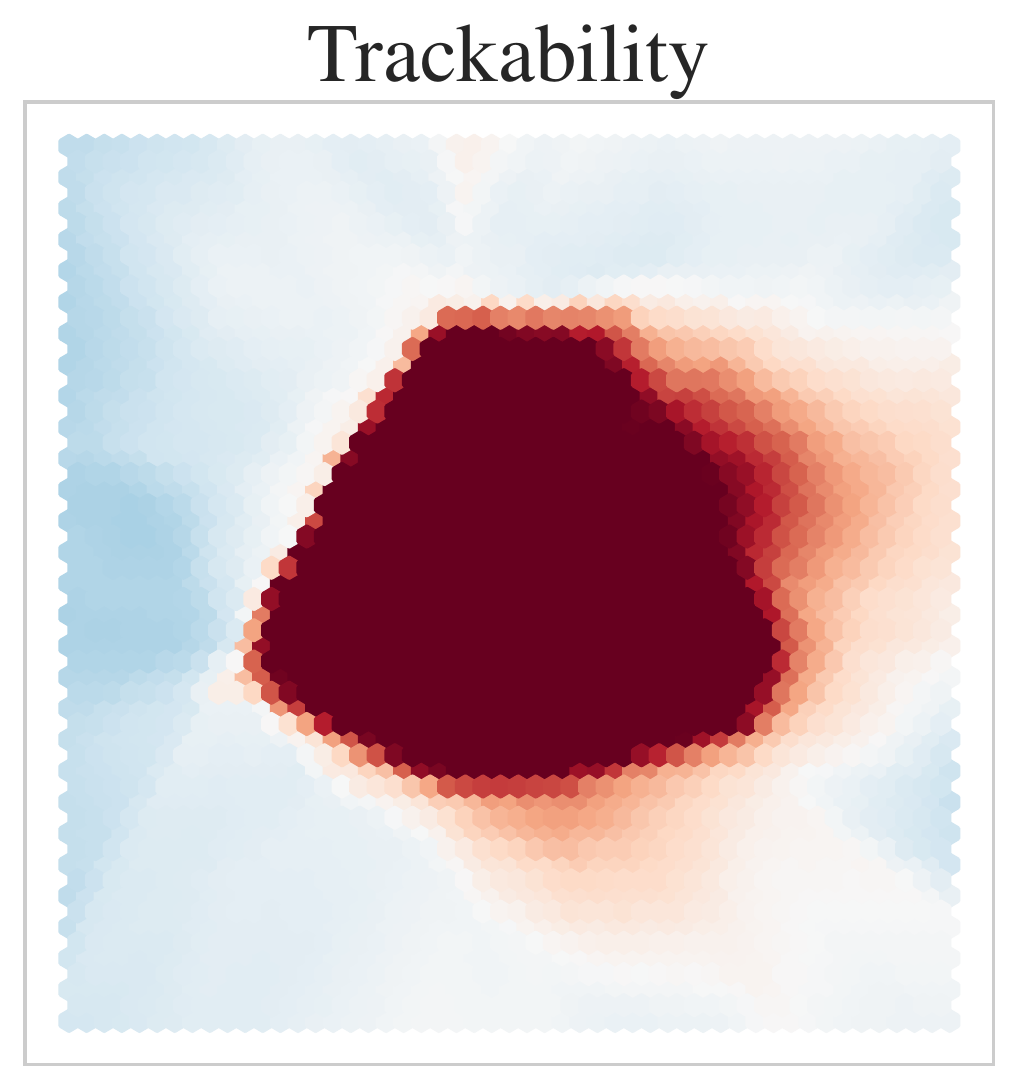}
        }\hfill
        \subfigure[][b]{
            \includegraphics[width=0.21\linewidth]{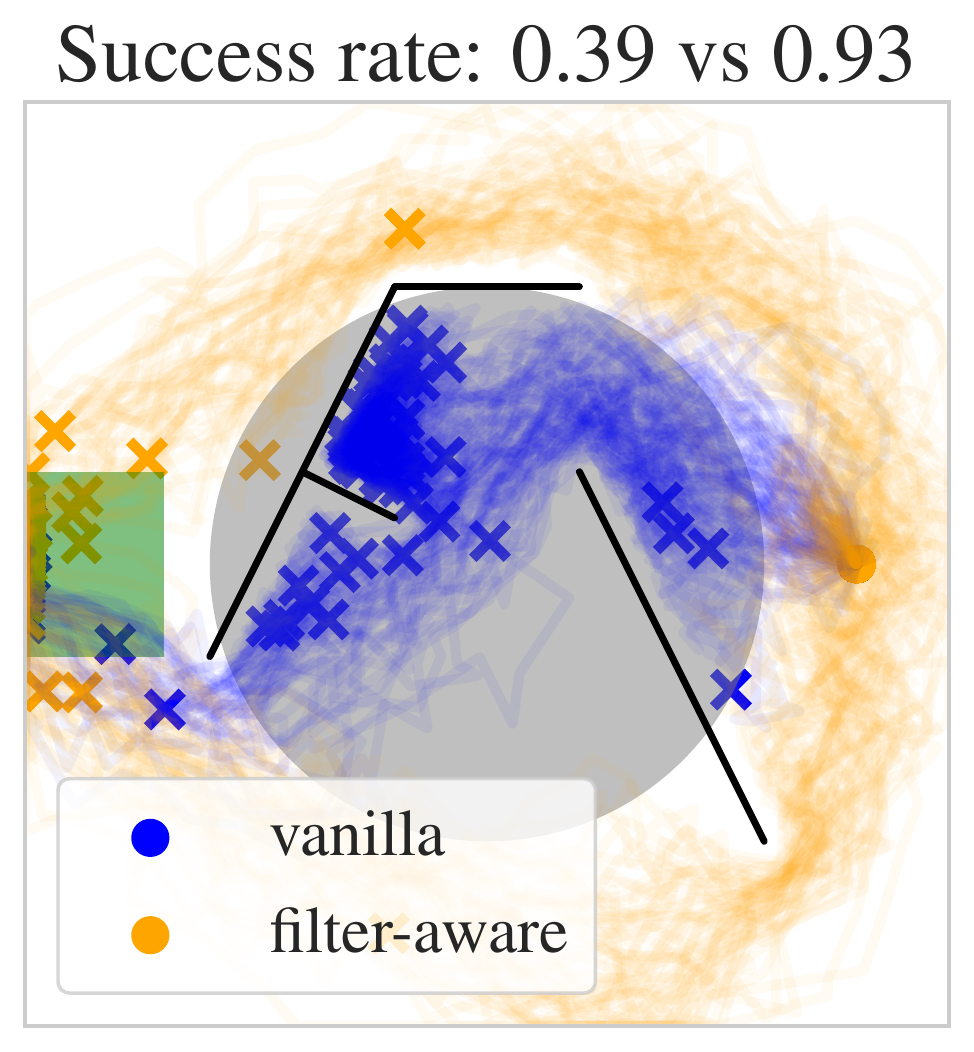}
        }\hfill
	\subfigure[][b]{
            \includegraphics[width=0.29\linewidth]{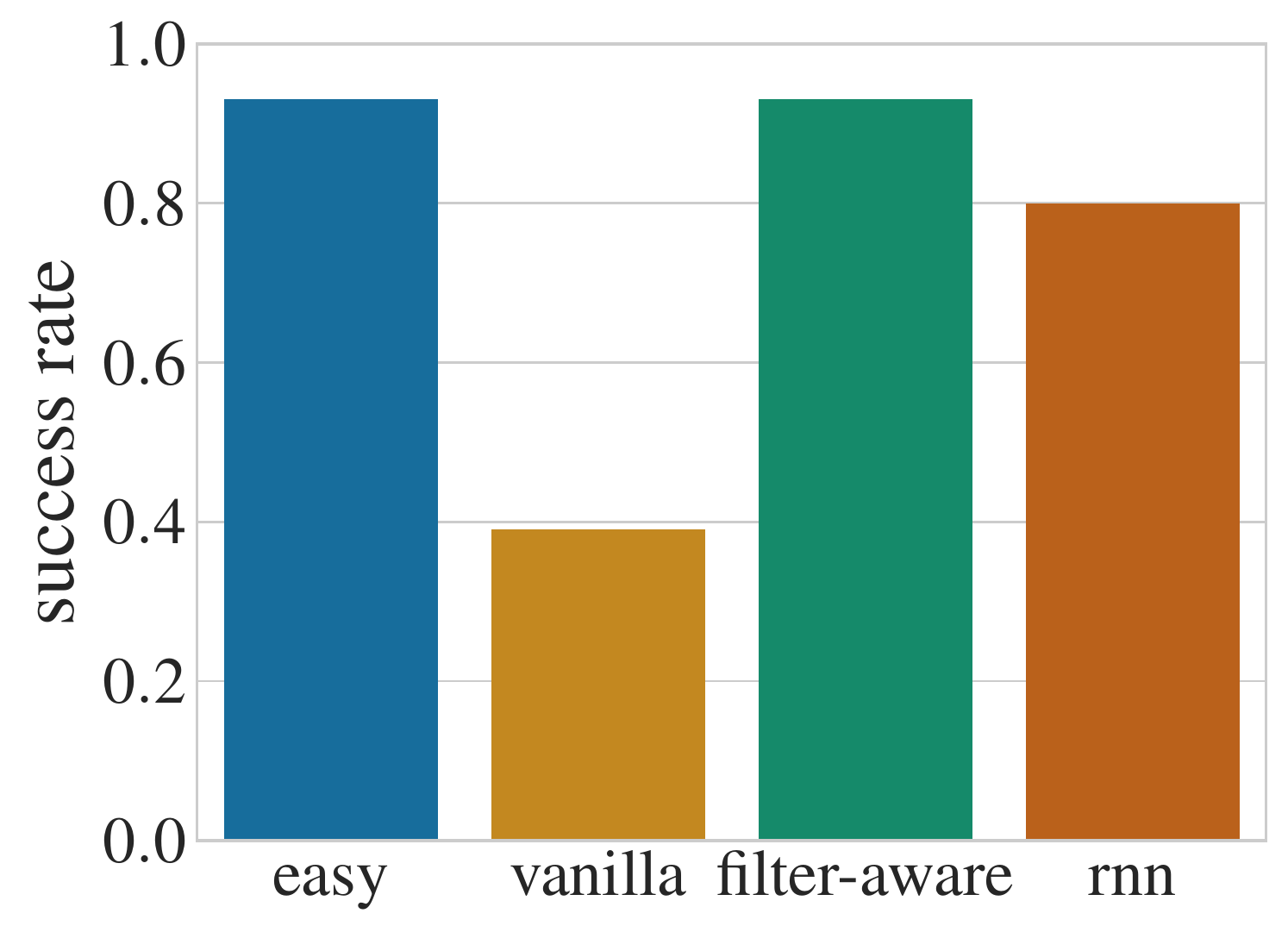}
        }
    }
    \vspace{-1em}
\end{figure}
We visualise this problem in \cref{fig:toy} (a).
We have a 2D agent that tries to reach the green box on the western side of the room.
The cost is the distance to the goal zone and the agent can observe its location with some additive Gaussian noise that is significantly higher inside a circle in the center (marked by the grey circle in \cref{fig:toy} (a)).
The agent can control its velocity subject to additive noise.

An easier problem, where the observation noise is the same everywhere and has a scale of $0.03$, can be solved by a model-predictive controller and a bootstrap particle filter with $93\%$ success, as shown in \cref{fig:toy} (d).
If we increase the observation noise to $1.0$ inside the grey circle, the controller fails, because MPC leads the agent inside the circle, where the particle filter is no longer able to keep track.
We can learn the trackability with a neural network, as outlined in \cref{sec:method}.
For that, we use 500 rollouts of the MPC policy with 30 time steps each.
We plot the network's output in \cref{fig:toy} (b), which reveals that the network can separate the problem area from the safe zones.
If we use the network to formulate constraints, the agent starts avoiding the grey circle and taking a slightly longer but safer route, as shown in \cref{fig:toy} (c).

We compare filter-aware MPC against vanilla MPC and an RNN policy in \cref{fig:toy} (d).
The RNN policy is trained by gradient descent on a differentiable implementation of this system to minimise the total cost within a horizon of 50.
We find that filter-aware MPC alone recovers the original success rate under this harder version of the problem, outperforming both vanilla MPC and the RNN, which can theoretically implement belief planning, as it has access to the full interaction history.

\subsection{Navigation in \vizdoom}
We now turn to a visual navigation problem based on the \vizdoom~simulator \citep{wydmuch2018vizdoom}.
Here we use coloured point-to-plane ICP \citep{chen1992object, steinbrucker2011real, audras2011real} as our state estimator since it is a well-established method for visual tracking.
\begin{figure}[t]
    \floatconts{fig:nav}{
        \caption{
            \vspace{-2em}
	    \vizdoom~Setup. The agent starts in one room and must reach the other. (a) Vanilla MPC picks the left corridor where observations are corrupted. (b) Sample observation. (c) Trackability function and filter-aware rollouts. (d) Vanilla MPC on an easy problem where both corridors are safe vs vanilla MPC on the harder problem vs filter-aware MPC on the harder problem, where the left corridor corrupts observations.
        }
    }%
    {
        \centering
        \subfigure[][b]{
            \includegraphics[width=0.23\linewidth]{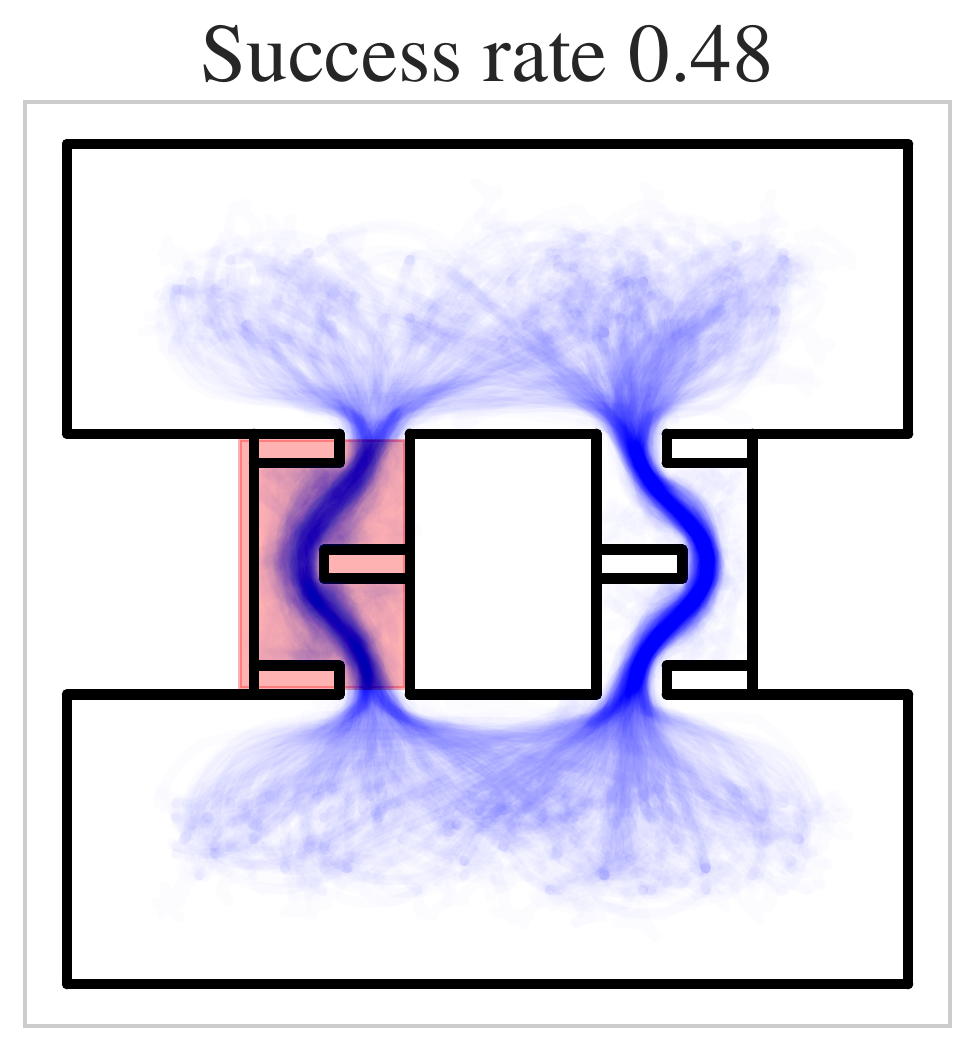}
        }\hfill
        \subfigure[][b]{
            \includegraphics[width=0.23\linewidth]{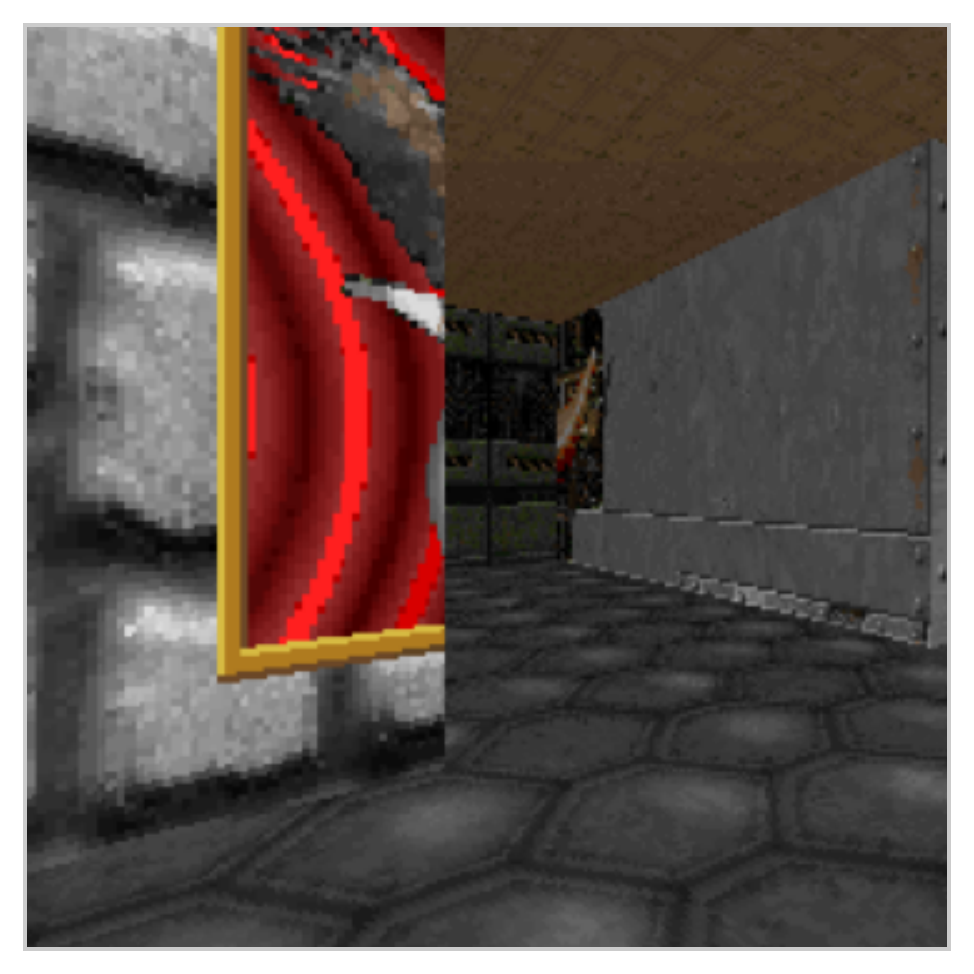}
        }\hfill
        \subfigure[][b]{
            \includegraphics[width=0.23\linewidth]{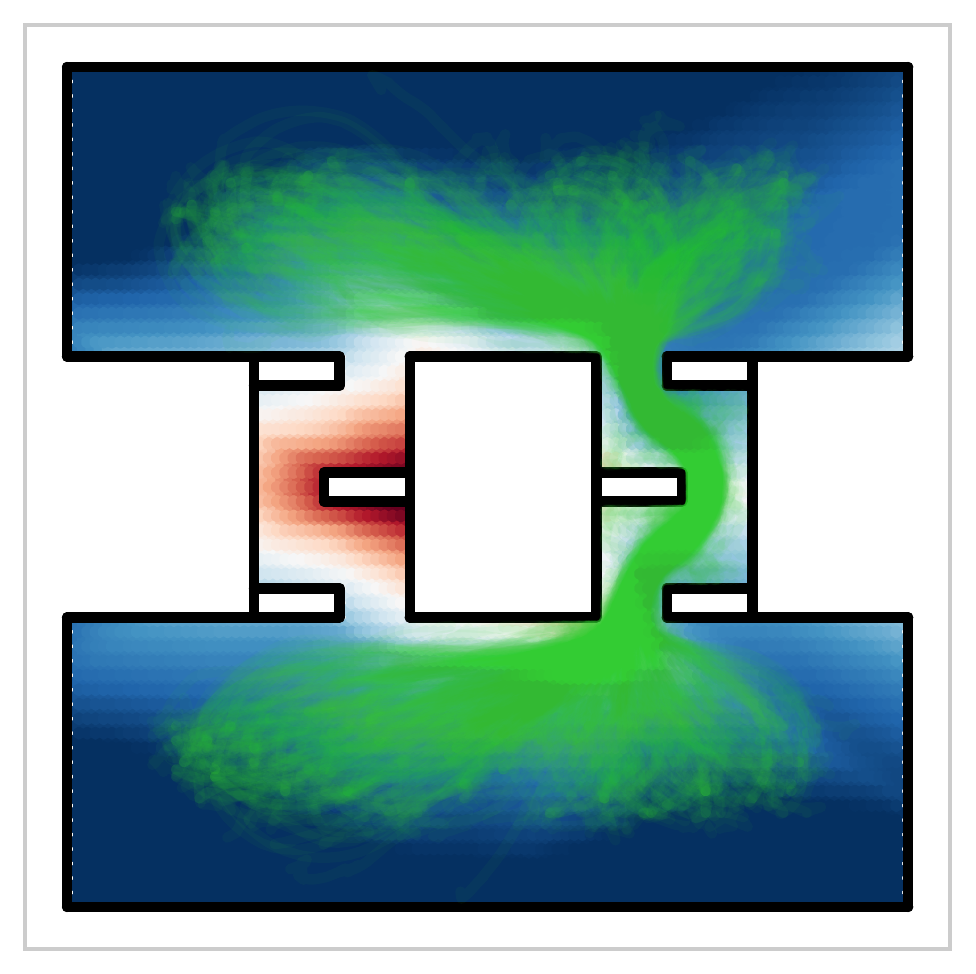}
        }\hfill
	\subfigure[][b]{
            \includegraphics[width=0.23\linewidth]{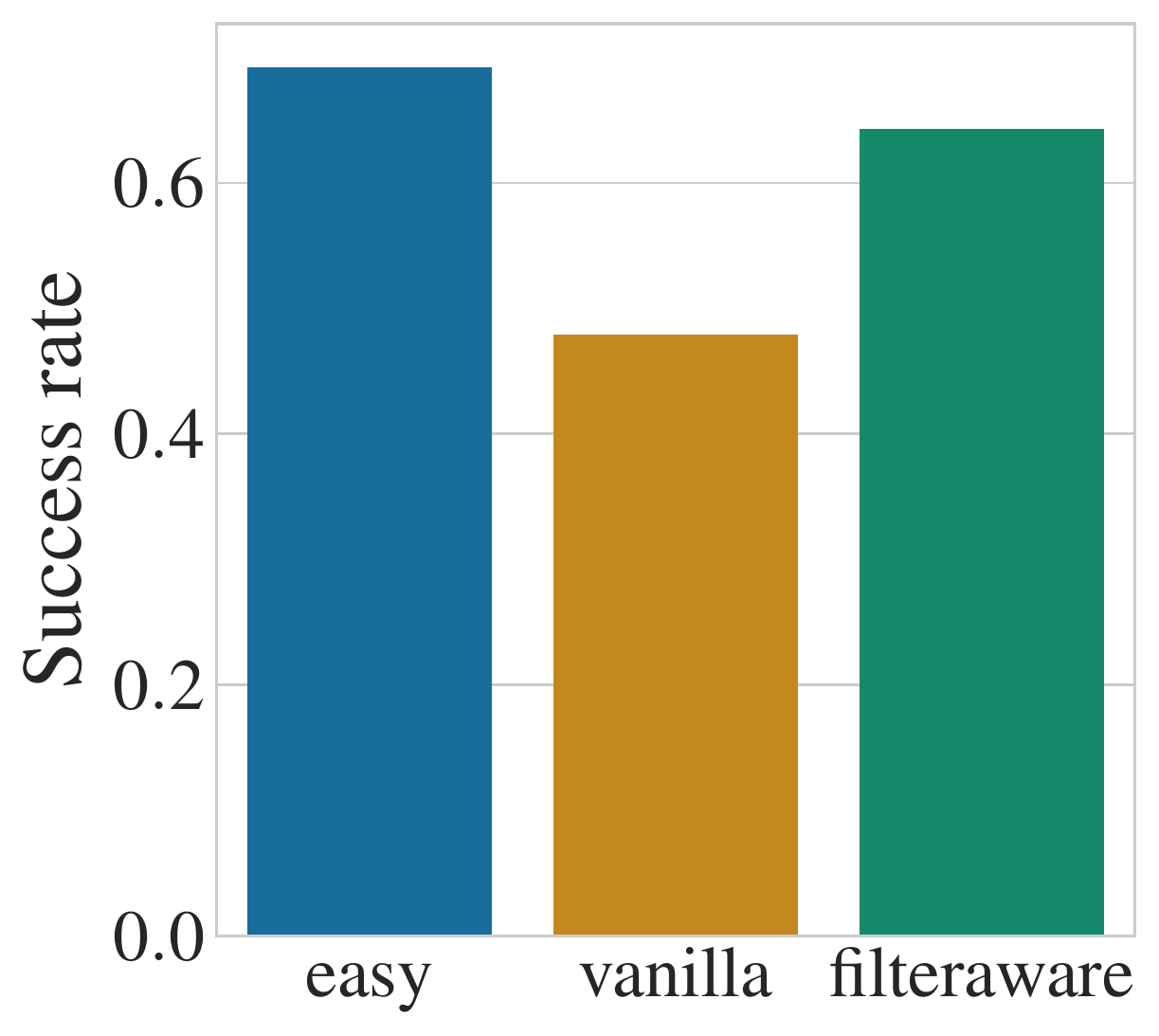}
        }
    }
    \vspace{-1em}
\end{figure}
The \vizdoom~environment has two rooms that are connected by two corridors.
The agent is placed in a random location in one room and must go to a random location in the other room.
It can see the world with an RGB-D camera and moves by picking a turning angle and a speed which is applied along its facing direction.
These controls are perturbed by noise, making the dynamics stochastic.
We show a sample RGB image in \cref{fig:nav} (b).

We make the problem harder by corrupting observations in the left corridor, analogous to the grey circle from our toy experiment.
This time, we zero-out the agent's RGB-D observation in the left corridor, meaning the agent can only rely on its transition model.
Our goal here is to simulate issues that plague visual tracking in the real world such as poor lighting conditions or reflective surfaces.
We show how navigation with vanilla MPC works under these conditions in \cref{fig:nav} (a).
The agent selects both corridors equally, even though the left corridor has poor observability, leading to a success rate of $48\%$.

Filter-aware MPC increases the success rate to $64\%$.
We learn the trackability using 900 rollouts of the MPC policy, having 200 time steps each.
We plot the learned trackability function and a set of filter-aware rollouts in \cref{fig:nav} (c).
The filter-aware controller avoids the left corridor in all cases.
We compare filter-aware MPC, vanilla MPC and vanilla-easy, which is vanilla MPC run on a setup where both corridors are safe in \cref{fig:nav} (d).
Filter-aware MPC outperforms vanilla MPC by $16\%$ when both are run on the same problem and approaches the performance of vanilla-easy, which achieves a success rate of $69\%$.

\subsection{Orbiting in a Realistic Environment}
The next experiment is designed to test filter-aware MPC under realistic conditions with naturally occurring features that are dangerous for the state estimator.
We use the \aithor~simulator \citep{ai2thor} which features models of realistic living spaces.
Real-life environments contain areas that are harder for a visual tracker to work with: blank walls, reflective surfaces, areas that cause sensor interference.
We experiment with this type of situation in a navigation-adjacent task, where a robot equipped with a visual tracker is trying to follow a fixed trajectory.

We show our test scene in \cref{fig:topdown} (a).
The green markers are landmarks that the agent must follow.
The agent receives RGB-D images that are passed to the same visual tracker as in the \vizdoom~experiments.
The agent can control its turning angle and speed with added noise.
Unlike in \vizdoom, the camera's facing angle can be controlled independently of its motion.
The cost on the agent's position and moving angle is defined to make it follow the landmarks.
As the cost does not depend on the camera's facing angle, the vanilla controller simply tries to keep the camera facing towards the agent's movement direction.
\begin{figure}[t]
    \floatconts{fig:topdown}{
        \caption{
            \vspace{-2em}
	    Orbit task. (a) Top-down view. The agent must follow the green landmarks. Blue is the average vanilla MPC trajectory, purple the filter-aware one, grey the ORB baseline. (b) Box plots for state estimation errors. (c) Box plots for task cost.
        }
    }%
    {
        \centering
        \subfigure[][b]{
            \includegraphics[width=0.28\linewidth]{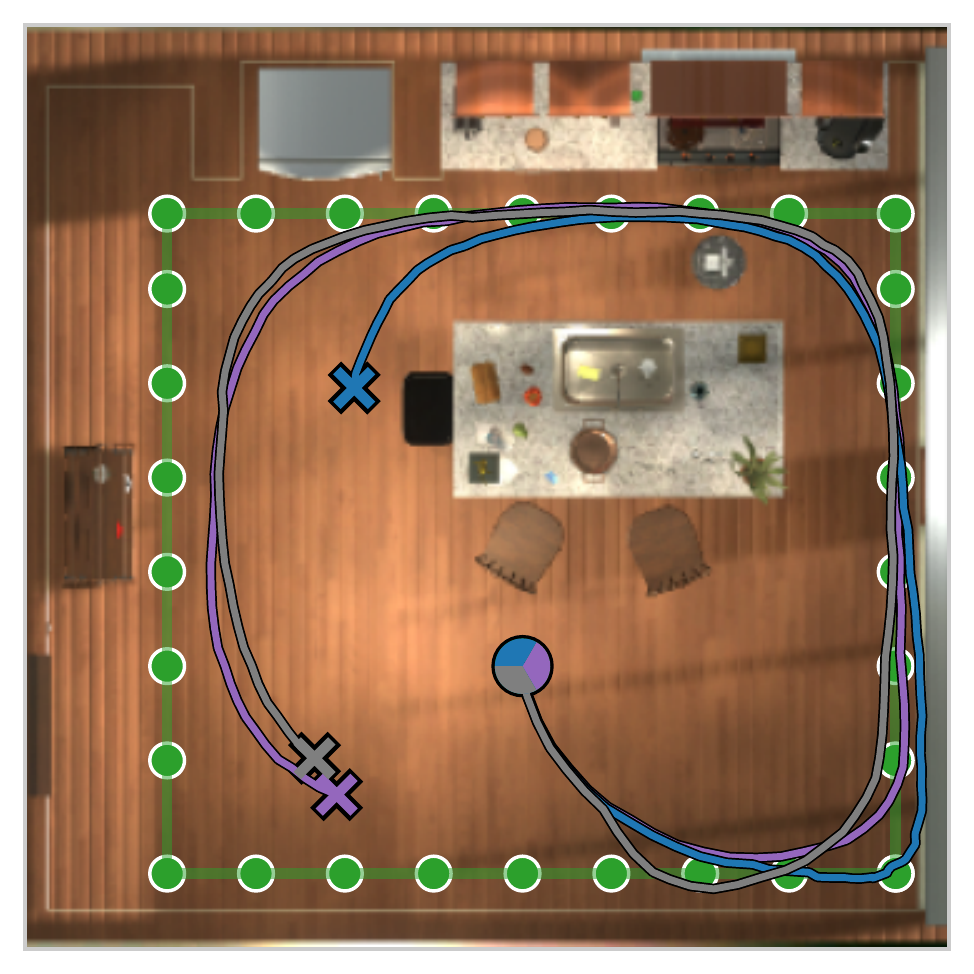}
        }\hfill
        \subfigure[][b]{
            \includegraphics[width=0.32\linewidth]{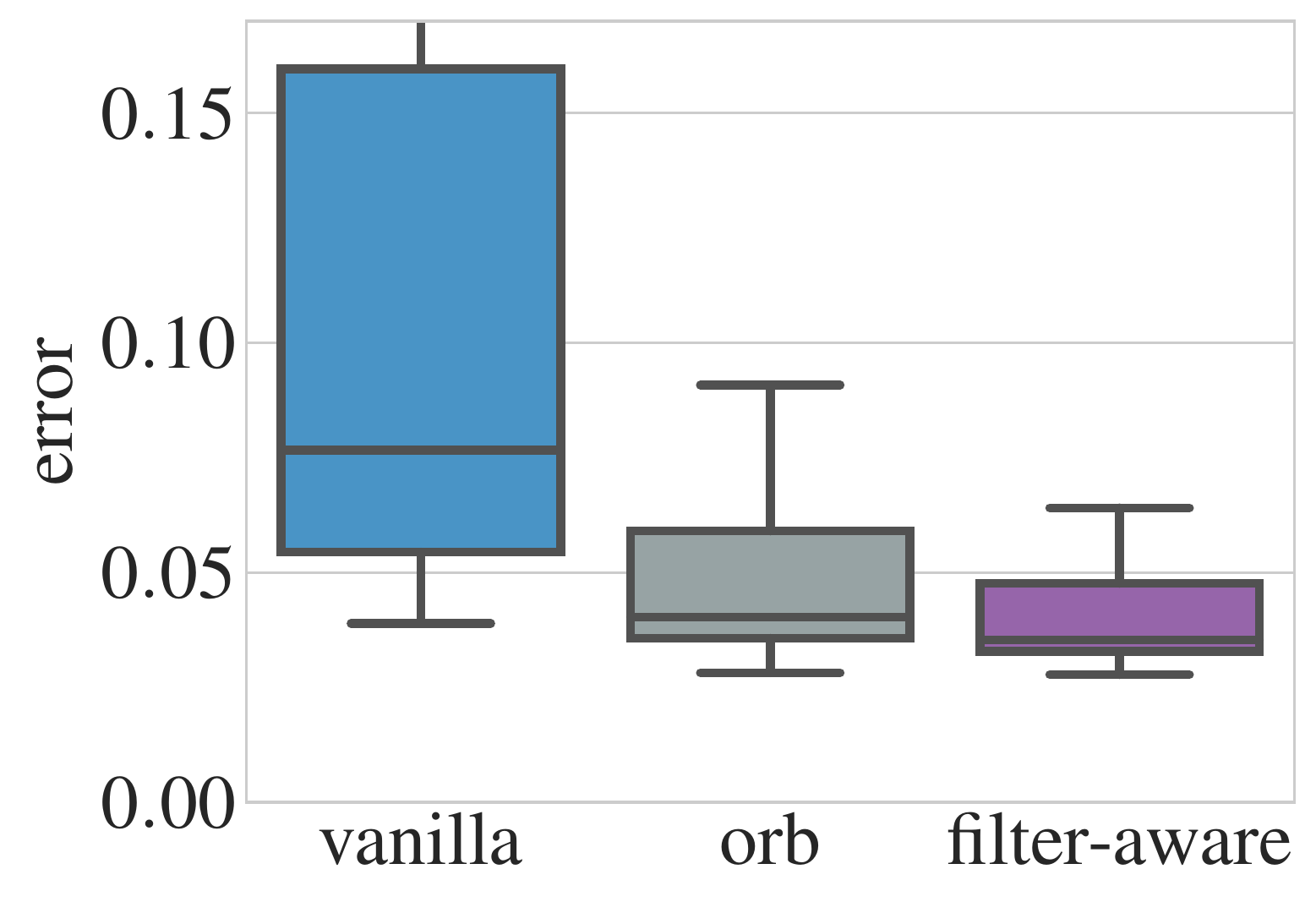}
        }\hfill
        \subfigure[][b]{
            \includegraphics[width=0.32\linewidth]{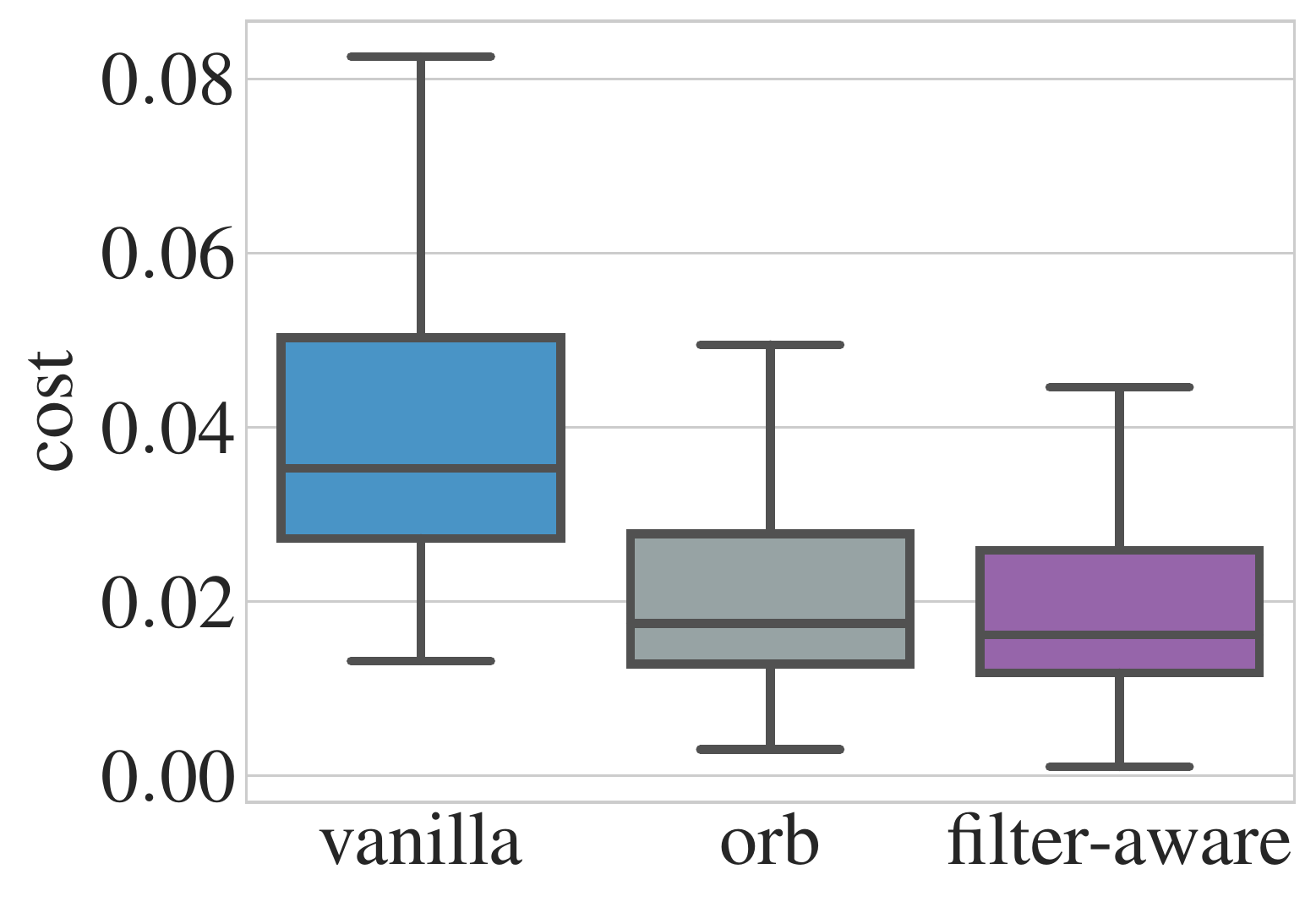}
        }
    }
    \vspace{-1em}
\end{figure}
\begin{figure}[t]
    \floatconts{fig:video}{
        \caption{
            \vspace{-2em}
	    Orbit task. (a) The naive strategy looks directly at the reflective surface, causing tracking failure. (b) Filter-aware MPC avoids looking at the reflective surface.
        }
    }%
    {
        \centering
        \subfigure[][b]{
            \includegraphics[width=\textwidth]{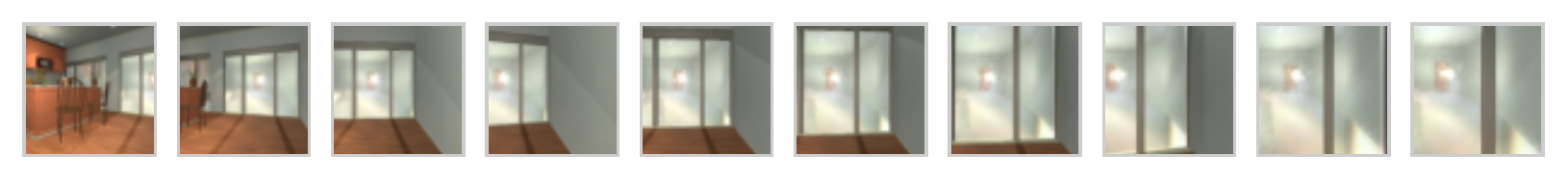}
        }\hfill
        \subfigure[][b]{
            \includegraphics[width=\textwidth]{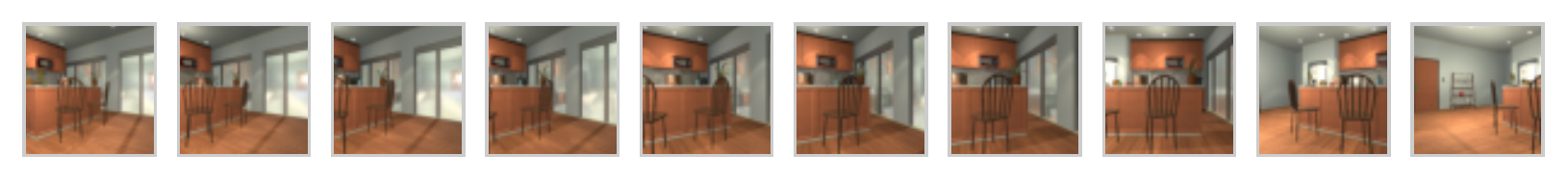}
        }
    }
    \vspace{-1em}
\end{figure}
The environment contains a large reflective surface in the eastern wall, which can throw off a visual tracker.
A filter-aware controller should be able to detect that looking at the reflective surface is dangerous for the tracker and look elsewhere, even while the agent is moving towards the reflective surface.
We find this to be the case, as shown in \cref{fig:video} (a--b).
Here, the trackability is learned from 4500 rollouts of the MPC policy with length 30.
While the naive controller looks where it moves, filter-aware MPC detects that doing so will cause the tracker to fail and instead looks towards the center of the room.
In practice, looking where you move would be necessary to avoid obstacles.
That can still be done with a filter-aware controller by using additional sensors, like a depth sensor or an additional camera for detecting obstacles.

We compare filter-aware MPC against two baselines: a) vanilla MPC b) an engineered baseline which uses ORB features \citep{rublee2011orb} as a proxy for trackability.
The ORB baseline learns a neural network to predict the number of ORB features that will be visible given a camera pose and uses that value as a constraint during control, analogous to filter-aware MPC.
We designed this baseline to resemble methods that focus on visual navigation \citep{falanga2018Pampc, rahman2020Uncertainty}.
The number of ORB features will be low when the camera is pointed at blank walls or the reflective surface on the eastern wall, and high when it is pointed towards the center of the room, which contains many strong visual landmarks.
Following areas densely covered by ORB features should therefore make it easier for the visual tracker to estimate the agent's state.

In \cref{fig:topdown} we draw a comparison between the three methods in terms of state estimation errors (b) and task cost (c) based on a set of 100 rollouts of length 200.
Both the ORB baseline and filter-aware MPC improve significantly over the vanilla strategy, with the filter-aware controller performing the best in terms of state estimation.
We show average rollouts for each method in \cref{fig:topdown} (a).
Both the ORB baseline and filter-aware MPC complete the course, with filter-aware having a slight edge.

\subsection{Planar Two-Link Arm with Occluded Regions}
In our final experiment we use a two-link robot arm that tries to reach random targets, based on the \brax~\citep{brax2021github} implementation of the classic reacher environment.

The agent can observe its joint angles and a vector pointing from the tip of the arm to the target.
To make state estimation more difficult, we set all observations to zero whenever the agent's arm enters the left-hand part of the work space (detected by checking the $x$-coordinate of a set of points on the arm).
The agent is never started inside this region and the targets are always on the right-hand side such that the agent can avoid losing track of its state if it is careful.
We add Gaussian noise to the controls before they are applied.
We use a bootstrap particle filter for state estimation.
We learn the trackability from 5000 rollouts of length 50 using the MPC policy.
\begin{figure}[t]
    \floatconts{fig:reacher}{
        \caption{
            \vspace{-2em}
	    Reacher task. (a) Box plots. \enquote{easy} is vanilla MPC used on an easier problem. (b) Top-down view. The circle shows a radius of $0.05$ around the target. Observations are missing beyond the dashed line. The colour indicates trackability. (c) A filteraware (green) vs a vanilla rollout (orange) in configuration space. The ellipse shows a radius of 0.05 around the target. Vanilla MPC enters the danger zone and loses track of itself.
        }
    }%
    {
        \centering
        \subfigure[][b]{
            \includegraphics[width=0.35\linewidth]{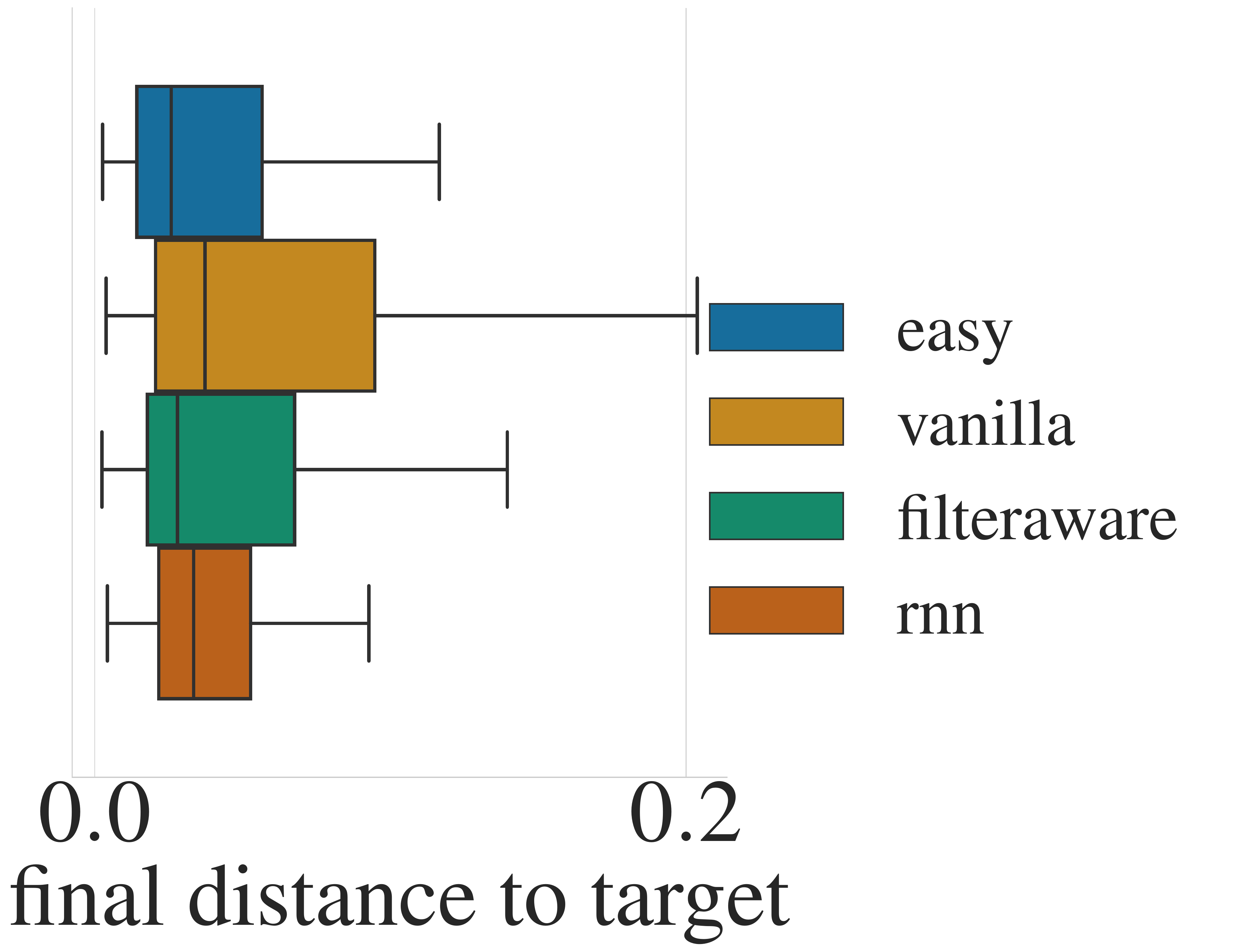}
        }\hfill
        \subfigure[][b]{
            \includegraphics[width=0.25\linewidth]{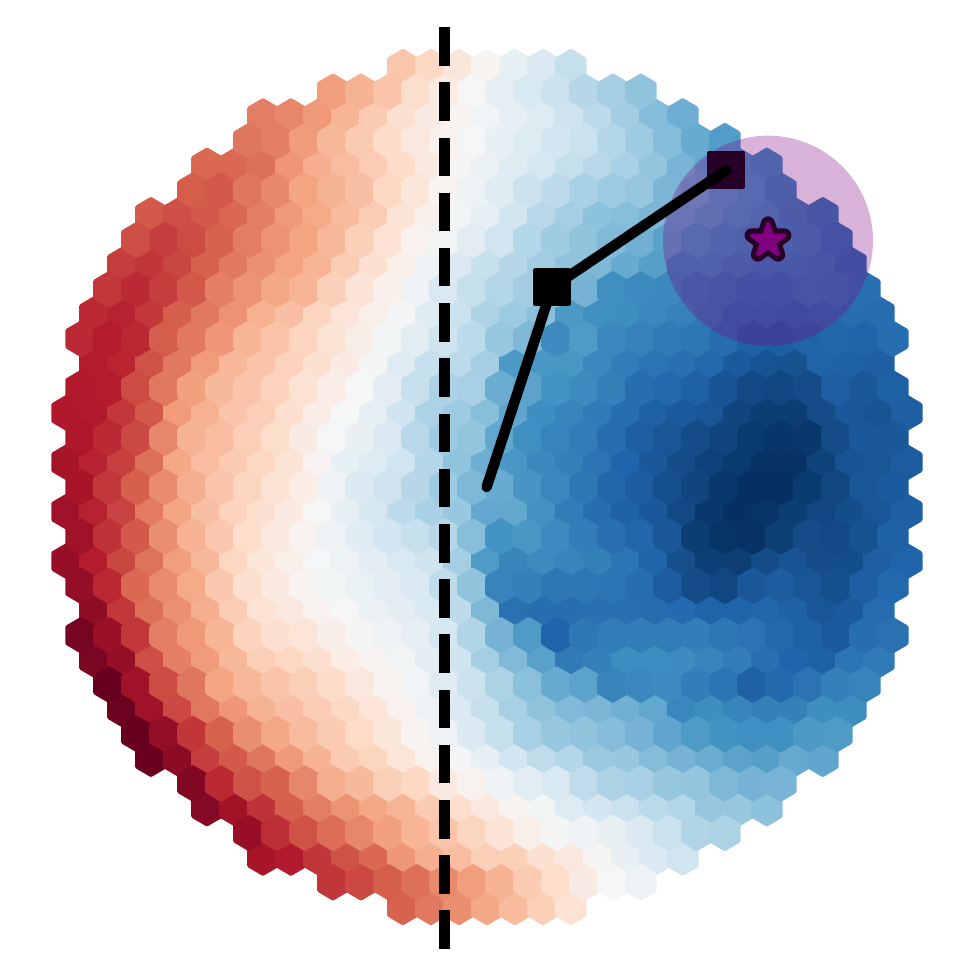}
        }\hfill
        \subfigure[][b]{
            \includegraphics[width=0.34\linewidth]{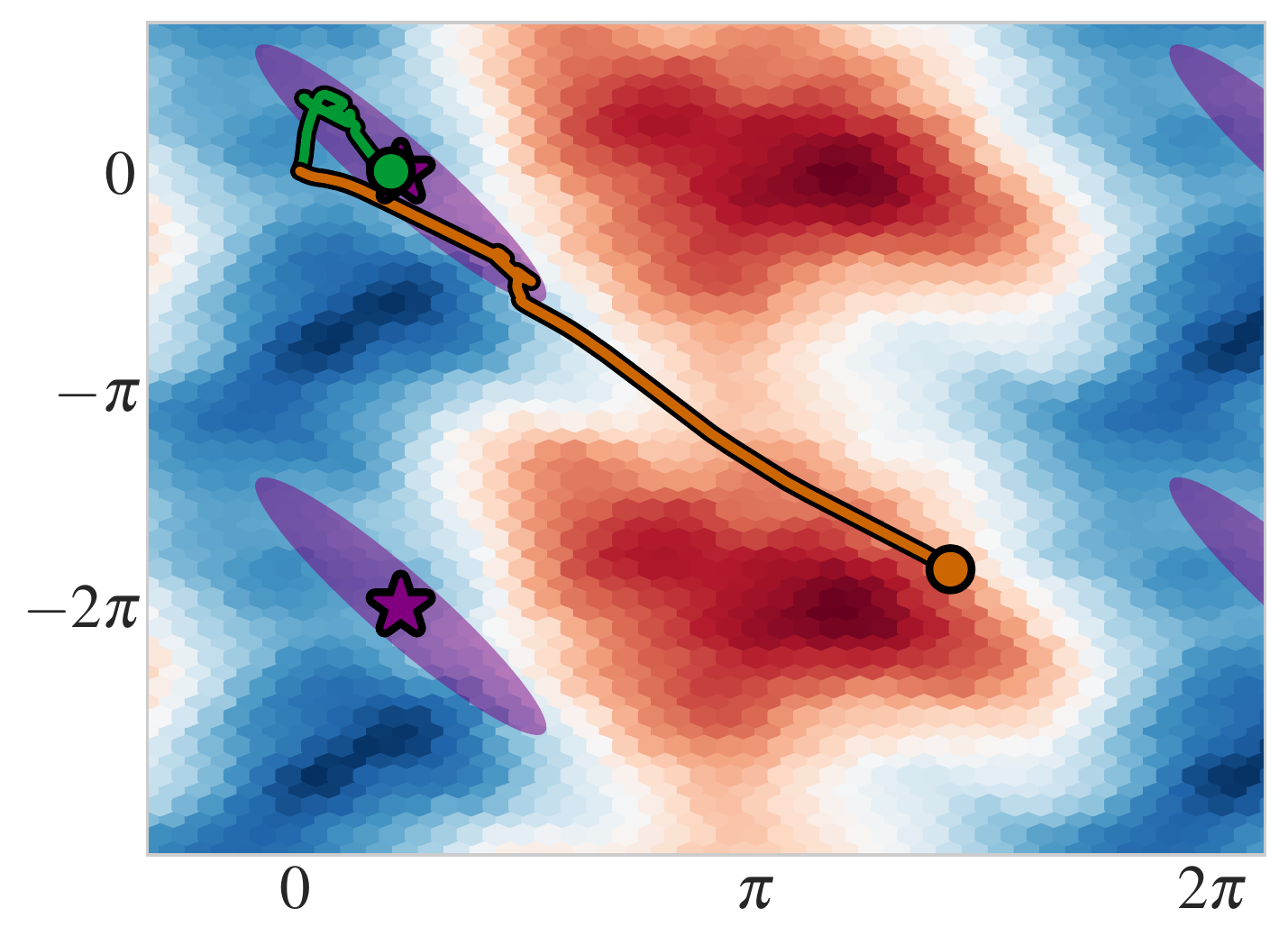}
        }
    }
    \vspace{-1em}
\end{figure}

We compare filter-aware MPC, vanilla MPC and an RNN policy in \cref{fig:reacher} (a).
The RNN policy follows the implementation of \citet{ni2021recurrent}.
Both the RNN policy and filter-aware MPC improve over vanilla MPC, though the RNN has an edge here.
We also present the performance of vanilla MPC in an easy setup where observations are available everywhere.
Note that the RNN policy performs better than vanilla MPC even when vanilla MPC is used on the easy problem.
This suggests that the performance gap between filter-aware MPC and the RNN might be due to the difference between MPC and amortised policy learning.
In other words, MPC itself appears to be at a general disadvantage compared to an RNN policy in this problem.

We can inspect the learned trackability function in \cref{fig:reacher} (b).
The network separates the blind zone from the safe area.
In \cref{fig:reacher} (c), we compare a filter-aware rollout and a regular one in the configuration space of the arm.
Regular MPC crosses into the blind zone and cannot recover.
\section{Limitations of Filter-Aware MPC}
Filter-aware MPC reduces state estimation errors by avoiding plans that are difficult to track.
We do this by placing constraints on the plan.
Any constrained controller can work worse than the unconstrained version if the constraints are too limiting.
We picked thresholds for trackability by hand in each experiment, relying on visualisations of trackability over the state space.
This is only possible in problems that can be visualised well in at most three dimensions.
In higher-dimensional settings the thresholds would have to be picked by hyper-parameter search.

The core assumption of our method is that we can learn trackability from rollouts.
We assumed that either the true environment model or a reasonably accurate approximate model is available, and used the true system in each of our experiments.
\section{Conclusion}
We introduced filter-aware MPC, an improvement over vanilla MPC for partially observable problems.
Filter-aware MPC constrains the planner such that it preserves a certain level of state estimation accuracy.
This bridges the gap between regular MPC, which only tries to reduce the cost without reasoning about future beliefs, and belief planning, which is often too expensive.
Filter-aware MPC depends on a measure of trackability which can be learned by a neural network, enabling filter-aware planning with little additional cost.
In experiments on increasingly realistic problems, we have tested filter-aware MPC and found that it consistently improves over vanilla MPC.
 
\bibliography{related}

\appendix
\section{Trackability Learning}\label{sec:tracklearn}

We use $\tdlambda$ to learn a trackability network.
Different papers have translated the $\tdlambda$-update rule into different neural network update schemes.
The one we used in this paper combines gradient descent with the parameter-averaging method used by \citet{fujimoto2018addressing}.

The objective that we minimise is:
\eq{
\mathcal{L}(\phi) = \sum_{i}^N (\phi(\state_1^i) -\track(\state_1^i;\lambda,\phi^\prime))^2,
}
with:
\eq{
\track(\state_1^i;\lambda,\phi^\prime) &= (1-\lambda)\sum_{k=1}^{T-1} \lambda^{k-1}\track_k(\state_1^i;\phi^\prime), &
\track_k(\state_1^i;\phi^\prime) &= \sum_{t=1}^k \discount^{t-1}\error_t^i + \discount^k\phi^\prime(\state_{k+1}^i).
}
Here, we introduced $\phi^\prime$, which is a different copy of $\phi$ that is only used to create targets for $\mathcal{L}(\phi)$.
The learning rule mixes gradient updates on $\mathcal{L}(\phi)$ with parameter-averaging updates for $\phi^\prime$:
\eq{\label{eq:update}
	\phi &\leftarrow \phi + \alpha\nabla\mathcal{L}(\phi) \\
	\phi^\prime&\leftarrow \eta*\phi^\prime + (1-\eta)*\phi.\numberthis
}
The parameter-averaging factor $\eta \in (0, 1)$ controls the speed at which the targets of $\phi$ will change.
We wrote a standard gradient descent update in \cref{eq:update} for simplicity.
In practice we use the adam optimiser \citep{kingma2017adam}.
\section{Approximating the Expected Future Tracking Error}

Filter-aware MPC approximates $\track(\state_t, \carry_t)$, which is the expected future tracking error with $\track(\state_t, \carryperf_{\state_t})$, which is the expected future tracking error {\it starting from a perfectly accurate initial state estimate}.
In our main text we have motivated this from a standpoint of practicality.
Working with $\carry_t$ would force us to predict this quantity into the future while planning.
Using $\carryperf_{\state_t}$ removes this need.

Here we would like to discuss another benefit of this approximation.
Since we are interested in further approximating trackability with a neural network, the generalisation error of this network would be critical for downstream performance.
If we were to try to learn $\track(\state_t, \carry_t)$, we would be approximating a function of both the system state $\state_t$ and the state estimator carry $\carry_t$.
Since the carry subsumes the history of environment interactions, accurately approximating this function would force us to train the neural network with many different interaction histories.
On the other hand, $\track(\state_t, \carryperf_{\state_t})$, the quantity we are actually interested in learning, is only a function of the system state $\state_t$ (since $\carryperf_{\state_t}$ is itself deterministically given by $\state_t$).
Thus, accurately approximating this function does not require exhaustively exploring the space of environment interactions.

At the same time, the fact that we learn trackability as a function of the state means that we can only reason about state estimation errors which can be modelled as a function of the state.
Our experiments feature environments where this is the case and state estimation errors result from local features of a region of space (e.g. due to the presence of a reflective surface).
As a counter-example, consider the {\it Heaven and Hell} problem by \citet{thrun1999Monte}.
Here, an agent has to reach either the north-western or south-eastern corner of a room, which is decided randomly without the agent's knowledge.
To find out its destination, the agent has to travel to the north-eastern corner and ask an oracle.
In Heaven and Hell, it is equally hard to estimate the location of the goal until the agent queries the oracle.
Once the agent has asked the oracle, state estimation is possible all across the room.
We cannot reason about the state estimation accuracy by looking at the state alone in Heaven and Hell, because it depends on whether the agent has interacted with the oracle already.
In other words, we need to know the interaction history.
\section{Computational Cost of Filter-Aware MPC}\label{sec:compcost}

Filter-aware MPC adds minimal computational cost on top of vanilla MPC.

\begin{table}
\centering
	\begin{tabular}{lr}
		\toprule
		\textbf{method} & \textbf{operations} \\
        \midrule
		state planning & J + h * (D + C) \\
		filter-aware planning & J + h * (D + C + T)  \\
		belief planning & J + h * (D + C + E + F) \\
		\bottomrule
\end{tabular}
\caption{\label{tab:compcost}Operations involved in planning. The letters denote evaluations of D - the dynamics model, J - the terminal cost, C - the stage cost, T - the trackability network, E - the emission model, F - the state estimator. The planning horizon is denoted by h.}
\end{table}

We give a loose overview in \cref{tab:compcost}.
Filter-aware planning has an additional cost of evaluating the trackability network in every stage of the planning horizon.
Full belief planning additionally requires evaluating the emission model and the state estimator.

A runtime comparison between state planning, filter-aware planning and belief planning would depend on the runtimes of the individual components.
For instance, if the dynamics D are much more expensive than the rest, we would expect all three to behave similarly.
On the other hand, if the trackability network is much more expensive to evaluate than the cost and the dynamics, filter-aware planning would be noticeably slower than state planning.
In environments involving high-dimensional observations such as images, the emission and state estimation steps would be much more expensive than the rest, making belief planning intractable.

We have computed runtimes for our implementation in each experiment and compare these in \cref{tab:runcost}.
First, we would note that our method works in real time in all environments except ViZDoom, where vanilla MPC also does not run in real-time.
We surmise that planning in ViZDoom is slowed down by several factors.
The first is the terminal cost evaluation.
Our terminal cost in ViZDoom is the length of the shortest path to the target, which is pre-computed for every location in state-space using dynamic programming and then stored in a 2D array.
During planning, we read from this array via bilinear interpolation using the agent's continuous 2D location.
This operation might slow down the execution.
Another potential source of issue is the collision modelling.
Here we have an inefficient implementation which simply checks for collisions against every obstacle, even ones that are infeasibly far away.

Similar to ViZDoom, we see that vanilla and filter-aware MPC share the same runtime for the reacher setup.
We assume that the planning is bottlenecked by the dynamics modelling through Brax in this environment.
In the toy setup and in AI2-THOR we find that filter-aware MPC runs slower by factors of $1.6$ and $1.75$ respectively.

\begin{table}
\centering
	\begin{tabular}{lcccc}
		\toprule
		\textbf{method} & \textbf{toy} & \textbf{vizdoom} & \textbf{ai2thor} & \textbf{reacher} \\
        \midrule
		vanilla & 192 Hz & 3 Hz & 28 Hz & 46 Hz \\
		filter-aware & 119 Hz & 3 Hz & 16 Hz & 45 Hz  \\
		\bottomrule
\end{tabular}
\caption{\label{tab:runcost}Runtimes in each environment.}
\end{table}

\section{A Note on Our Practical Implementation}\label{sec:optimise}

In our implementation we used a random search algorithm for simplicity and speed.
Though this procedure gives no formal guarantees for constraint satisfaction, we found that it worked well in our experiments.
We would also point out that individual constraint violations in our setting might not be as critical in others, since visiting a state where tracking is difficult will not necessarily result in direct state estimation failure.
This is contrast to safety-critical constraints such as obstacle avoidance.

Similarly, we have no guarantees of finding the optimal plan, and are limited by the set of random plans that we sample.
It is possible to replace our crude optimisation algorithm with more sophisticated approaches with better guarantees.
Filter-aware MPC does not impose any restrictions on vanilla MPC aside from the constraints.
As such, any optimisation algorithm that is applicable to MPC in a certain setting can also be used for filter-aware MPC, as long as it can handle constraints.
\section{MPC and Terminal Costs}

We use terminal costs for two of our experiments: the toy scenario and \vizdoom.
These problems have sparse costs, which makes it necessary to use a terminal cost, which is then also the driving factor when planning.
This is because the cost signal has the same value for every plan that is not close to the target.
The standard MPC objective with a terminal cost is:
\eq{
\mathbb{E}_{\state_t|\obs_{1:t}, \control_{1:t-1}} \left [ \sum_{k=t}^{t+K} \discount^{k-t} c(\state_t, \control_t) + \discount^{K+1} {\apxtotalcost}(\state_K) \right ].
\numberthis
}
In practice we have found that we get better performance by using the $\lambda$-return as our objective.
This is given by the following equations:
\eq{
\obj(\state_1^i;\lambda) &= (1-\lambda)\sum_{k=1}^{T-1} \lambda^{k-1}\hatobj_k(\state_1^i), &
\hatobj_k(\state_1^i) &= \sum_{t=1}^k \discount^{t-1}\cost_t^i + \discount^k\apxtotalcost(\state_{k+1}^i).
}
This is a weighted average over MPC objectives with different planning horizons.
We believe it worked better than the regular MPC objective due to the sparseness of the cost and the stochastic dynamics.
When the cost is sparse, the terminal cost is the only factor that guides planning, until the agent is close enough to the target to reach it within the planning horizon.
This is an issue under stochastic dynamics because the transition error will build up towards the end of the planning horizon.
The end of the planning horizon will have the highest prediction error, though it will also be the only place where we check the terminal cost.

We can reduce the noise by shortening the planning horizon, but that in turn can hurt the performance if the terminal cost is not a perfect approximation of the true future total cost.
This is the case in the \vizdoom~environment, where the terminal cost is just the geodesic distance from a location to the target, assuming the agent can move in any direction that is not blocked by a wall.
The geodesic distance doesn't take into account that the agent can only move along its facing direction and can only change that direction by a limited amount in each time step.
In some places the agent needs time to turn and face a better direction.
A longer planning horizon is necessary to make sure the agent has enough time to reorient itself and then also move for long enough to reach a lower terminal cost.

The $\lambda$-return is a middle-ground between a long planning horizon that has noisier planning and a low horizon that allows less complex maneuvers.
This only applies to the two environments with sparse costs however, and in the rest of the environments we used the regular MPC objective without terminal costs.
\section{Visual Tracking}\label{sec:vistrack}
The state estimator we used for \vizdoom~and \aithor~is based on colored point-to-plane ICP \citep{chen1992object, steinbrucker2011real, audras2011real}.
Since MPC assumes some knowledge of the transition dynamics, it makes sense to merge this into the tracker as well.
We modify the basic algorithm slightly, adding the squared distance from the transition model's prediction as another loss term into the objective.
This was previously done by \citet{kayalibay2022tracking} and section 4.2 of their paper summarises the approach.
Note that we use the true simulator for emissions while their paper uses a world model.
\section{Relationship to Belief Planning}\label{sec:belief}
Belief planning optimally trades-off cost reduction and information gathering.
This comes at the cost of reasoning about beliefs.
Exact solutions are often only possible in heavily restricted setups \citep{Littman96algorithmsfor, KaelblingLC98, cassandra1998incremental}.
Thus, much of the research on belief planning focuses on efficient approximations, e.g. by representing the value function as a maximum over a set of vectors \citep{parr1995approximating} or by reducing the continuous space of beliefs to a discrete set of representative points \citep{pineau2003point}.

Continuous state spaces present a further challenge to belief planning, as there is no trivial solution to representing beliefs over a continuous state space.
Here, particle filters are a popular approximation \citep{thrun1999Monte}.
Indeed particle filters have enabled a range of tree search algorithms capable of handling continuous state, observation and action spaces \citep{Sunberg2017Online, sunberg2017Value}.

Gaussian beliefs are another common choice, used with extended or unscented Kalman filters.
Several works present belief planning approaches for the Gaussian case \citep{todorov2005stochastic}.
\citet{platt2010belief} do belief planning with EKFs by reducing the space of possible observations to the mode of the emission distribution.
\citet{vanDenBerg2021Efficient} seek to approximate the composition of transition, emission and state estimation in an EKF.
\citet{rafieisakhaei2017tLQG} improve LQG planning by planning around a previously computed trajectory and optimising for state estimation accuracy.
A notable work by \citet{rahman2020Uncertainty} allows application to vision-based observations by approximating the posterior belief given an observation using the number of features in the observation.

Compared to these approaches, our work will be suboptimal in the sense that it does not plan in the space of beliefs and is not capable finding the optimal balance between state estimation and cost reduction.
On the other hand, our approach has several benefits.
Filter-aware MPC plans in state space, which is generally simpler than planning in belief space for several reasons.
First, the dynamics in state space are often simpler than those in belief space.
Further, predicting forward in state space is computationally cheaper, since we are not forced to reason about the emission distribution or the state estimate update while planning.
Belief planning in continuous state spaces also requires a tractable and rich enough representation of the belief for meaningful planning such as particle filters or Gaussian beliefs.
We are not limited to such representations, and can work with any black-box state estimator.
Indeed, our experiments feature both particle filters (in the toy setup and the reacher experiments) and a visual tracker that uses point-to-plane ICP (in ViZDoom and AI2-THOR), which only provides a maximum a-posteriori estimate of the state.

Another benefit of our approach is its applicability to high-dimensional observations such as images.
By comparison, the tree-search algorithms presented in \citep{Sunberg2017Online} and \citep{sunberg2017Value} are tested on much lower-dimensional environments.
The work of \citet{rahman2020Uncertainty} is applicable to image observations, but they rely on an approximation of the state posterior which only works with feature-based visual tracking.
We side-step issues related to the observation dimensionality by only learning the error of the state estimator and carrying out the rest of our computations in state space.
\section{Experiment Details}

\subsection{Toy setup}

\paragraph{Problem setting.}
The system state is the 2D location of the point.
The observations are the state with additive Gaussian noise.
The controls are 2D velocities, with a speed limit of $0.05$.
The cost is 0 when the point is inside the goal area, 1 otherwise.
Walls block the agent's movement.
Each control is perturbed by additive Gaussian noise with a scale of $0.03$ before being used.
The observation noise is $0.03$ outside of the dark zone and $1.0$ inside.
The dark zone is centered at $(0.5, 0.5)$ and has a radius of $0.3$.
The environment is contained in the square area $[0, 1]^2$.

\paragraph{Dynamics equations.}
Let $\state \in \mathbb{R}^2$ be a state, $\obs \in \mathbb{R}^2$ and observation and $\control \in \mathbb{R}^2$ a control with $\lVert \control \rVert_2 \le 0.05$.
The system dynamics are:
\eq{
\tilde \control_t &= \control_t + \noise_t \hspace{2.0em}\text{ with }\noise_t \sim \mathcal{N}(0, 0.03), \\
\state_{t+1} &= \begin{cases}\state_t + \tilde \control_t & \text{no collision} \\ \state_t - 0.01\tilde \control_t / (\lVert \tilde\control_t \rVert_2 + 0.0001) & \text{collision}\end{cases}, \\
\obs_t &= \state_t + \obsnoise_t\hspace{2.5em}\text{ with }\obsnoise_t \sim \mathcal{N}(0, \sigma_{\noise}), \\
\sigma_\noise &= \begin{cases}0.03 & \lVert\state_t - \begin{bmatrix}0.5 \\ 0.5\end{bmatrix}\rVert_2 \ge 0.3 \\ 1.0 & \text{otherwise}\end{cases}.
}
Here, "collision" means the line segment that starts at $\state_t$ and ends at $\state_t + \tilde \control_t$ intersects one of the obstacles.
In this case we take a small step in the opposite direction by normalising $\tilde \control_t$ to have unit length and scaling it up to $0.01$.
The additive constant $0.0001$ is to avoid a division by zero.
\paragraph{State estimator.}
We use a bootstrap particle filter.
The proposal distribution is a Gaussian.
Its mean is given by the transition function's prediction and its scale is the same as the transition noise: $0.03$.
We use 512 particles for estimating the initial state and 128 particles after that.

\paragraph{Terminal cost.}
The sparse cost requires using a terminal cost.
We use the geodesic distance to the target area as a terminal cost.
For filter-aware MPC, we re-calculate geodesic distances by respecting the constraint.
If a location violates the constraints, we don't allow the shortest path to lead through that location.
This is the same as treating the constrained areas as obstacles while calculating shortest distances.

\paragraph{Random search hyper-parameters.}
The proposal distribution samples a random control from a uniform distribution for the first time step and repeats it for the rest of the horizon.
We take 100 candidates with a horizon of 10.
For each plan we estimate the expected future total cost with 50 Monte Carlo samples.
We execute the first control only, before re-estimating the state and re-planning.
The MPC objective is the $\lambda$-return with $\lambda=1.0$ and $\discount=1.0$.

\paragraph{Trackability data.}
We collect rollout for trackability learning by using MPC on the true system.
Here, the planning horizon is set to 5.
We take 500 rollouts with 30 steps.
The initial location is sampled uniformly over the space.

\paragraph{Trackability learning.}

We estimate trackability with $\tdlambda$, setting $\lambda=0.95$ and $\discount=0.8$.
We divide the training rollouts into chunks of length 5.
The tracking error is defined as the weighted sum of the squared error of each particle, weighted by its particle weight.
The parameter-averaging factor (see \cref{sec:tracklearn}) is set to $0.995$ and we use 5000 gradient updates.
The learning rate is set to $0.001$.
We use use 128 hidden units and two hidden layers with relu activations.
The batch size is 512.

\paragraph{Comparative study.}

We compare filter-aware MPC and regular MPC using 100 rollout of length 50.
The {\it easy} setting uses an emission noise of $0.03$ everywhere, omitting the dark zone.
For filter-aware MPC, the constraint threshold $\delta$ is set to $0.6$.

\paragraph{RNN Baseline.}

The RNN baseline is trained by gradient descent to minimise the total cost in a horizon of length 50.
The initial state is sampled uniformly to give the agent access to experience from all over the environment.
The cost is modified to be the length of the shortest path to the goal, i.e. the geodesic distance to the goal.
We found this to be crucial for solving the task with a parametric policy.
We precompute geodesic distances to the goal using a uniform grid and index the grid by discretising the current location of the agent.
We define a custom gradient for this operation using finite differencing.

\subsection{\vizdoom}

\paragraph{Problem setting.}
The system state is the 2D location and 1D yaw of the agent.
The observations are RGB-D images.
The controls are the turning angle and forward velocity, with a maximum angle of $6^\circ$ and a maximum speed of $0.015$.
The cost is 0 when the agent is within a radius of $0.1$ of the goal, 1 otherwise.
The controls are perturbed by clipped Gaussian noise.
The clipping ensures that noise does not flip the direction of the control.
If the agent tries to turn left by $2^\circ$, the noise is clipped to be in $[-2, \infty)$, such that noise cannot cause it to turn right instead of left.
Likewise, if the agent tries to move forward with a speed of $0.01$, the speed noise is clipped to be in $[-0.01, \infty]$ such that it cannot move backward instead of forward.
When the agent enters the left corridor, every pixel in the RGB-D observation is set to zero.
The environment is contained in the square area $[0, 1]^2$.

\paragraph{Dynamics equations.}
Let $\alpha \in [0, 360^\circ]$ be an orientation and $l \in \mathbb{R}^2$ a location.
The system state is a concatenation of these: $\state = [\alpha, l]^T$.
Let $\dot \alpha \in [-6^\circ, 6^\circ]$ be a turning angle and $s \in [-0.015, 0.015]$ a forward speed.
The control is a concatenation of these: $\control = [\dot \alpha, s]^T$.
The dynamics are then:
\eq{
	\alpha_{t+1} &= \alpha_t + \text{sign}(\dot \alpha_t)\max(0, \mid \dot \alpha_t \mid + \epsilon_t^\alpha) \text{ with } \epsilon_t^\alpha \sim \mathcal{N}(0, 2^\circ), \\
	l_{t+1} &= l_t + \begin{bmatrix}\cos(\alpha_{t+1}) \\ \sin(\alpha_{t+1})\end{bmatrix}\max(0, s_t + \epsilon_t^s) \text{ with } \epsilon_t^s \sim \mathcal{N}(0, 0.0035).
}
The way in which noise is injected into the dynamics was previously used by \citet{kayalibay2022tracking}.
Note that these dynamics are further subject to collision handling with the environment.
Here, we only allow moving from $l_t$ to $l_{t+1}$ if the line segment connecting these points does not intersect any wall.
If it does intersect a wall, we only allow moving up to the intersection point.
The observations $\obs$ are RGB-D images created by the ViZDoom simulator.

\paragraph{State estimator.}
We use colored point-to-plane ICP \citep{chen1992object, steinbrucker2011real, audras2011real} for state estimation (see \cref{sec:vistrack}).

\paragraph{Terminal cost.}
The sparse cost requires using a terminal cost.
We use the geodesic distance to the goal as a terminal cost.
For filter-aware MPC, we re-calculate geodesic distances by respecting the constraint.
If a location violates the constraints, we don't allow the shortest path to lead through that location.
This is the same as treating the constrained areas as obstacles while calculating shortest distances.

\paragraph{Random search hyper-parameters.}
The proposal distribution samples a random direction between left and right and turns in that direction for a random number of time steps using the maximum turning angle.
We also sample a random time index to start moving forward in and then move forward with the maximum speed for a random number of time steps.
We take 200 candidates with a horizon of 20.
For each plan we estimate the expected future total cost with a single Monte Carlo sample.
We execute the three controls of each plan, before re-planning.
The MPC objective is the $\lambda$-return with $\lambda=1.0$ and $\discount=1.0$.

\paragraph{Trackability data.}
We collect rollout for trackability learning by using MPC on the true system.
We take 900 rollouts with 200 steps.

\paragraph{Trackability learning.}

We estimate trackability with $\tdlambda$, setting $\lambda=0.95$ and $\discount=0.95$.
We divide the training rollouts into chunks of length 5.
We only accept a chunk into the training set if the tracking error at the second time step (i.e. after using the first control) is less than $0.99$.
The tracking error is defined as the squared distance between the agent's true and inferred locations.
The network's input is the agent's location only (i.e. we disregard the orientation component).
The parameter-averaging factor (see \cref{sec:tracklearn}) is set to $0.995$ and we use 10000 gradient updates.
The learning rate is set to $0.0001$.
We use use 128 hidden units and two hidden layers with relu activations.
The batch size is 512.

\paragraph{Comparative study.}

We compare filter-aware MPC and regular MPC using 90 rollout of length 200.
The {\it easy} setting uses the normal, uncorrupted RGB-D image in both corridors.
For filter-aware MPC, the constraint threshold $\delta$ is set to $3.5$.

\subsection{\aithor}

\paragraph{Problem setting.}
The state is the 2D location, 1D robot facing angle (yaw), and 1D camera angle relative to that.
The observations are RGB-D images.
Controls are 2D velocity, robot turning angle, camera turning angle.
The maximum speed is $0.04$ and the maximum turning angles are $10^\circ$.
The cost is designed to make the agent follow a loop, which is specified by a list of landmarks.
Given the agent's location $l_t$ at time $t$, we find the closest landmark $p^1_t$ and the next landmark from the list $p^2_t$.
We define a local target $p^*_t = p^1_t + 0.55 * (p^2_t - p^1_t)$.
The final cost is then:

\eq{
c(\state_t, \control_t, \state_{t+1}) = \lVert p^*_{t+1} - l_{t+1}\rVert + 10 * \langle \nu_{t+1}, p^2_{t+1} - p^1_{t+1} \rangle + 10 * \langle p^2_{t+1} - p^1_{t+1}, p^*_{t+1} - p^*_t  \rangle,
}

where $\nu_{t+1}$ is the velocity component of the control $\control_{t+1}$ and $\langle \cdot, \cdot \rangle$ is the scalar product.
The robot is initialised at (0.0, -2.5) facing the southern wall.
We use the scene "FloorPlan10" from the set of \aithor~environments.
We add zero-centered Gaussian noise with a scale of $0.02$ to the velocity controls.

\paragraph{Dynamics equations.}
Let $\alpha,\psi \in [0, 360^\circ]$ be orientations and $l \in \mathbb{R}^2$ be a location.
The state concatenates these: $\state = [\alpha, \psi, l]^T$.
Let $\dot \alpha, \dot \psi \in [-10^\circ, 10^\circ]$ be turning angles and $\nu \in \mathbb{R}^2$ a velocity vector with $\lVert \nu \rVert_2 \le 0.04$.
The control concatenates these: $\control=[\dot \alpha, \dot \psi, \nu]^T$.
Then, the dynamics are:
\eq{
	\alpha_{t+1} &= \alpha_t + \dot \alpha_t, \\
	\psi_{t+1} &= \psi_t + \dot \psi_t, \\
	l_{t+1} &= l_t + \nu_t + \noise_t\hspace{2em}\text{ with } \noise_t \sim \mathcal{N}(0, 0.02). \\
}
The observations $\obs$ are RGB-D images produced by the AI2-THOR simulator.
For that we teleport the camera to the 3D location $[l_t, 0.9]^T$ and yaw angle $\alpha_t + \psi_t$, where $0.9$ is the constant elevation of the camera.
Note that these dynamics are subject to collision handling done by AI2-THOR.

\paragraph{State estimator.}
We use the same state estimator as in the \vizdoom~experiments.
We assume that the initial state of the system is known.

\paragraph{Terminal cost.}
Since the cost is not sparse, no terminal cost was necessary in this environment.

\paragraph{Random search hyper-parameters.}
We use a combination of two optimisers for filter-aware MPC.
The first plans the controls for moving the robot: the velocity and turning angle.
The second takes the movement controls as given and only plans the camera angle.
Only the second controller looks at the constraints.
The first planner is implemented the same way as in the \vizdoom~experiments.
The second planner's proposal distribution first picks a random camera angle within the set of angles that can be reached in the planning horizon.
Then it finds the individual turning angles at every time step that are required to reach that camera angle and maintain it.
Both planners use the same hyper-parameters: 5000 candidate plans, a horizon of 5, 1 Monte Carlo sample per plan and a replanning interval of 3 time steps.
For regular MPC we only use the first planner and set the relative camera angle to zero, so that the camera is always facing the direction of movement.

\paragraph{Trackability data.}
We collect training data by using regular MPC on the true system, taking 4500 rollouts of length 30.
For this procedure we place the agent at a random location and facing angle.

\paragraph{Trackability learning.}
We estimate trackability with $\tdlambda$, setting $\lambda=0.95$ and $\discount=0.95$.
We do not further divide the training rollouts into subsequences, and use the original rollouts themselves.
We only accept a rollout into the training set if the tracking error at the second time step (i.e. after using the first control) is less than $0.99$.
The tracking error is defined as the squared distance between the agent's true and inferred locations.
The network's input is the agent's true location and absolute camera angle (i.e. robot facing angle plus the relative camera angle) converted to cosine-sine representation.
The parameter-averaging factor (see \cref{sec:tracklearn}) is set to $0.995$ and we use 10000 gradient updates.
The learning rate is set to $0.001$.
We use use 128 hidden units and two hidden layers with relu activations.
The batch size is 512.

\paragraph{Comparative study.}
We compare filter-aware and regular MPC with 200 rollouts of length 100.
The constraint threshold $\delta$ is set to -10.0, making sure the constraint is always active.
This is an appropriate choice here because we can optimise for safe trackability by turning the camera alone which can be done without impeding the movement of the robot.

\paragraph{ORB Baseline.}
The ORB baseline trains a neural network to predict the number of ORB features that will be visible under a given camera pose.
Here, we use the same rollouts that were used to train the trackability critic.
For each image, we detect ORB features with OpenCV \citep{opencv_library} and compute their number.
We record the camera's yaw and 2D location in a dataset, along with the number of features that were detected.
We train a neural network to map the camera pose to the feature count.
The feature count is normalised by a division by 500, the maximum number of features that is allowed in the feature detection phase.
The rest of the algorithm is identical to filter-aware MPC, with the trackability critic replaced by the ORB network.

\subsection{Reacher}

\paragraph{Problem setting.}
The state is the 52-dimensional physics state used by \brax~(referred to as {\it QP} in the \brax~documentation).
The observations are joint angles and the vector from the tip of the arm to the target.
The controls are joint actuation signals for each joint.
The cost is the absolute distance between the tip of the arm and the target.
Whenever this distance drops below $0.05$, the agent gets a bonus cost of $-100$ for one time step.
If the distance rises above $0.05$ again, this bonus is canceled out by a one-time penalty of $100$.
The initial state of the arm is as defined by the \brax~implementation of the reacher.
The target sampling is modified to make sure that targets are never in the dark zone, where observations are corrupted.
We inject noise into the dynamics by perturbing the controls with zero-centered Gaussian noise with a scale of $0.1$.
Whenever the x-coordinate of any point on the arm is less than $-0.02$ (detected by checking 20 equally-spaced points on the arm), the observations are replaced by zero.

\paragraph{State estimator.}
We use a bootstrap particle filter with Gaussian distributions placed on the transition and emission functions.
The transition scale is $0.005$ and the emission scale is $0.001$.
We assume the initial state is known, and use 100 particles for all future time steps.
Though the state-space if $52$-dimensional, many of the components are static and need to have specific values for accurate physics simulation.
We set these static components to their ground-truth values and keep them fixed.

\paragraph{Terminal cost.}
Since the cost is not sparse, no terminal cost was necessary.

\paragraph{Random search hyper-parameters.}
The proposal distribution samples a random control for each joint and then applies that control for a random duration and from a random start.
Some movements require both joints to be actuated at the same time, either with the same signal or with different signals (e.g. one joint turns left while the other turns right).
To make sure we get enough samples where this is the case, we use same random time interval for both links in one third of the candidates and use both the same random interval and the same random control signal in another third.
The remaining third samples different random intervals and controls for each joint.
We use 300 candidates and a planning horizon of $22$.
We take 5 MC-samples per plan and use the first 3 controls of every plan before replanning.

\paragraph{Trackability data.}
The training data is generated by using regular MPC on the true system.
We sample 5000 rollouts of length 50.
The initial arm position is random and uniform across the state space (including the dark zone) to ensure a good coverage.

\paragraph{Trackability learning.}
We estimate trackability with $\tdlambda$, setting $\lambda=0.95$ and $\discount=0.95$.
We do not further divide the training rollouts into subsequences, and use the original rollouts themselves.
The tracking error is defined as the weighted sum of the squared error of each particle, weighted by its particle weight, as in the toy experiment.
The network's input is the true joint angles converted to cosine-sine representation.
The parameter-averaging factor (see \cref{sec:tracklearn}) is set to $0.995$ and we use 20000 gradient updates.
The learning rate is set to $0.00001$.
We use use 128 hidden units and two hidden layers with relu activations.
The batch size is 512.

\paragraph{Comparative study.}

We compare filter-aware MPC and regular MPC with 100 rollouts of length 200.
The constraint threshold $\delta$ is set to $9.0$ for filter-aware MPC.

\paragraph{RNN Baseline.}
The RNN baseline follows the design of \citet{ni2021recurrent}.
We use the same model that was used in their \enquote{standard POMDP} experiments.

\end{document}